%% file: root.tex
\documentclass[runningheads]{llncs}

\usepackage{preamble}

\begin{document}

\title{From Robot Execution Traces to Monitoring Graphs: Combining VLMs and LLMs for Interpretable Skill Monitoring}
\titlerunning{Combining VLMs and LLMs for Interpretable Skill Monitoring}

\author{Alex Mitrevski\Letter\inst{1}\orcidID{0000-0003-3591-3160} \and
Karinne Ramirez-Amaro\inst{1}\orcidID{0000-0003-4600-1722} \and
Emmanuel Dean\inst{2}\orcidID{0000-0002-1155-7088}}
\authorrunning{A. Mitrevski, K. Ramirez-Amaro, and E. Dean}

\institute{Division for Systems and Control, Department of Electrical Engineering, Chalmers University of Technology, Gothenburg, Sweden
\email{\{alemitr,karinne\}@chalmers.se}\\
\and
ABB Robotics, V{\"a}ster\r{a}s, Sweden\\
\email{emmanuel.dean-leon@se.abb.com}}

\maketitle


\begin{abstract}
    Semantic representations of robot behaviour, such as those using logical representations, are commonly used for task description, planning, and execution monitoring; however, the process of executing individual skills, now usually done with learning-based skill representations, is typically opaque.
    This is a challenge when failures occur during execution, as it can be difficult to understand what the robot was actually trying to achieve.
    In this paper, we present a skill execution monitoring system whose objective is the \emph{extraction of interpretable, robot-agnostic execution monitors} in the form of labelled execution graphs.
    Our proposed framework starts with image observations from nominal execution trajectories of a given skill (such as picking up a spoon or pushing a cup); using open-world language models, we extract execution knowledge from these nominal trajectories and then summarise it as a graph of execution states, which can be used for online monitoring.
    Concretely, our system:
    (i) generates \emph{language narrations of individual execution steps} with respect to a vocabulary of predicates,
    (ii) combines the step-by-step narrations of each episode into an \emph{episode execution summary}, and then merges individual episode summaries into a \emph{skill execution summary}, which we refer to as a \emph{semantic envelope}, and
    (iii) converts the semantic envelope to a \emph{labelled monitoring graph} of execution states.
    We evaluate our framework on a set of five typical everyday skills (involving picking, placing, and pushing objects) with two different robots.
    A quantitative comparison with ground-truth descriptions of these skills shows high similarity of the generated episode summaries (cosine similarity generally higher than $0.85$); we also show statistically significant improvements over baseline summaries extracted with minimal or no concrete evidence from execution episodes.
    Qualitatively, we also demonstrate the potential of using the graphs for online execution monitoring, the importance of their interpretability in this context, and the ability to generalise the graphs between task objects and robots.
    Overall, our study indicates the benefit of providing detailed, guided information about the robot execution process when using open-world models for automatic skill monitor extraction, as well as the feasibility of embedding such monitors into a robot's online execution process.
    An associated website for this paper can be found at \href{https://alex-mitrevski.com/robot-semantic-envelopes/}{https://alex-mitrevski.com/robot-semantic-envelopes/}.
    \keywords{Robot execution monitoring \and Robot interpretability}
\end{abstract}


\input{subfiles/introduction}
\input{subfiles/related-work}
\input{subfiles/methodology}
\input{subfiles/evaluation}
\input{subfiles/discussion}


\section*{Declaration of AI-assisted technologies in the manuscript preparation process}

This manuscript is fully human-written; no generative AI tools were used during its preparation.


\section*{Acknowledgment}

This work was supported in part by the Swedish Research Council under Grant 2025-06377.


\section*{CRediT author contribution statement}

\textbf{Alex Mitrevski}: Conceptualization, Investigation, Methodology, Resources, Software, Validation, Visualization, Writing-original draft, Writing-review.
\textbf{Karinne Ramirez-Amaro}: Funding acquisition, Supervision, Writing–review.
\textbf{Emmanuel Dean}: Funding acquisition, Supervision, Writing–review.

\bibliographystyle{splncs04}
\bibliography{references}


\input{subfiles/appendix}

\end{document}

%% file: subfiles/introduction.tex
\section{Introduction}
\label{sec:introduction}

\begin{figure}[t]
    \centering
    \includegraphics[width=\linewidth]{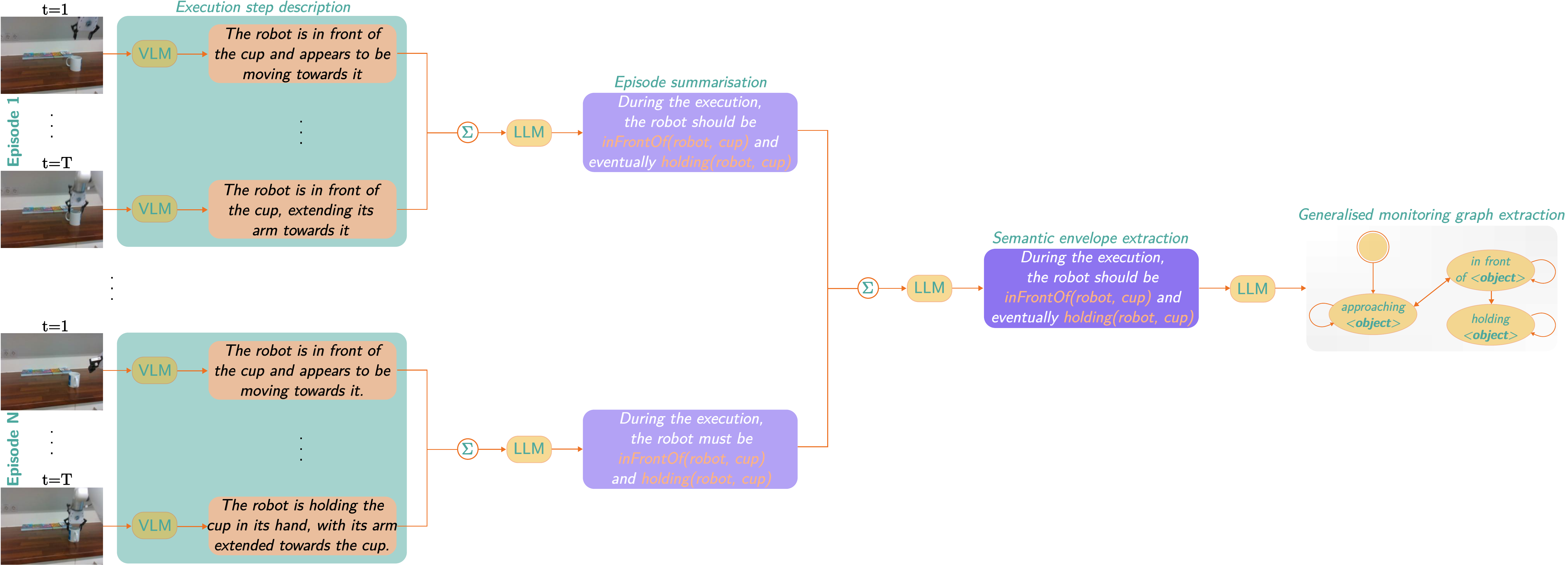}
    \caption{Overview of our method. We use a vision-language model to annotate execution steps in skill episodes; a language model then summarises the annotations into episode summaries and then condenses these into a semantic execution description (a \emph{semantic envelope}). This description is then converted into a lifted graph of execution states that can be used in an execution monitoring process. The lifted representation removes dependencies on specific object identities, potentially enabling the reuse of monitoring graphs across objects and robot embodiments, provided that the underlying semantic execution stages remain applicable and can be reliably recognised.}
    \label{fig:overview}
\end{figure}

Robots used in everyday applications need the ability to perform a variety of skills in different contexts, for instance carefully grasping an egg or plugging in a cable into a socket.
While performing such skills, a robot may experience \emph{execution failures}, such as knocking the egg over while moving towards it or missing the socket due to perceptual inaccuracies.
To be able to prevent such failures before they occur, it can be useful to know \emph{what} the robot is actually trying to achieve, not only at the overall task level, but also at the lower skill execution level; this can enable the robot to perform suitable preemptive actions or engage a human collaborator if necessary.
At the task level, semantic representations, typically based on qualitative logical models, already provide a solid basis that guides the understanding of a robot's objectives; a comparable semantic representation at the skill level would be a natural complement that ultimately leads to a more interpretable robot execution process.

The current predominant paradigm to drive a robot's behaviour is based on learning-based skills, as these provide a solid basis for flexible execution and lifelong behavioural adaptation \cite{kroemer2021}.
Particularly neural network-based policies trained at scale, for instance as foundation models, represent the current state-of-the-art of robot manipulation \cite{firoozi2025,openvla2024,octo2024,openx2024,pi02024,gupta2024,liang2023,cui2023}.
To use such models effectively, careful and detailed prompting is typically necessary \cite{liang2023,cui2023,wu2023,qi2026}, and simulations are sometimes also used to verify the outcomes of the generated behaviour \cite{gupta2024}.
Nevertheless, without finetuning, learning-based skills are typically unreliable on out-of-distribution tasks, and, when failures occur, it can be difficult to understand the underlying reasons.
This is primarily because any semantic knowledge about the executed skill is hidden in the learned policy representation.

Monitoring and recovery \cite{liu2023,agia2024,argenziano2025,sliwowski2025,santhanam2026} can mitigate the problem of failures, but such methods are not always associated with semantic descriptions or operate at a task-level granularity.
Explicit semantic representations are often used for task learning \cite{aein2019,ramirezamaro2017} and, in the form of linear temporal logic (LTL) descriptions, for task monitoring \cite{zudaire2021,quartey2025}; such representations, for instance in the form of logical descriptions and causal models, have also been shown to be useful for failure diagnosis and correction \cite{mitrevski2023,diehl2023}.
There is, however, a lack of studies that attempt to semantically describe the skill execution process itself, even though such descriptions can be beneficial for execution monitoring, explainability, and for more directed modification of a robot's execution behaviour.
Although existing approaches, such as \cite{sliwowski2025,zhou2025}, investigate execution monitoring and automatic extraction of task constraints, comparatively less attention has been devoted to constructing explicit, interpretable representations of intermediate skill execution stages directly from nominal visual execution trajectories.
Semantic descriptions of individual skill execution stages can provide an interpretable basis for monitoring robot behaviour and identifying deviations from expected execution patterns.
Such representations could subsequently support predictive failure detection and corrective interventions (although these capabilities are beyond the scope of this study).

In this paper, we study the problem of whether \emph{semantic skill execution descriptions} can be extracted by \emph{automatically analysing robot skill execution episodes}.
Our proposed method, illustrated in Fig. \ref{fig:overview}, generates semantic annotations (in the form of natural language captions) of observations collected during nominal execution (which can be seen a story describing the complete execution), and then combines the step-by-step annotations into a summary of the execution process.
Concretely, we use an open-world vision-language model (VLM) to generate annotations based on a predefined vocabulary (which is expressed as a set of lifted predicates), and then generate execution summaries using a large language model (LLM), either in natural language or in the form of LTL sentences.
We then convert the extracted summary of the execution process, which we refer to as a \emph{semantic envelope}, into a labelled execution graph, whose objective is to be used for skill execution monitoring.

Through this, we study the following concrete research questions:
(RQ1) Can pretrained, open-world VLMs and LLMs be effectively combined to extract meaningful semantic scene descriptions during robot execution in terms of conditions that a robot should fulfill during the execution?
(RQ2) Does the inclusion of semantic observational evidence improve upon implicit prior execution knowledge that is already encoded in pretrained LLMs?
(RQ3) Are existing models better at summarising observations in the form of natural language or temporal logic statements (in terms of semantic similarities to ground-truth summaries)?
(RQ4) Are extracted semantic scene descriptions useful for execution monitoring?

The main contributions of this work include:
(i) an end-to-end skill execution monitoring framework that describes robot execution episodes semantically and generates execution monitoring graphs based on the descriptions,
(ii) a detailed quantitative analysis of the correctness of generated semantic descriptions, showing statistically significant improvements with respect to baseline description generation methods,
(iii) a demonstration that the execution graphs resulting from the semantic envelopes can be useful for execution monitoring, as well as that they can be generalised to other objects and between robots, and
(iv) an open-source implementation of our method.\footnote{The implementation is available at \url{https://github.com/alex-mitrevski/robot-semantic-envelope-generation}}

To understand the scope of our study better, it is also important to discuss a few aspects that we do not consider in this work:
(a) Firstly, our concrete objective is to improve the interpretability of learned skill execution by associating observed robot behaviour with explicit semantic execution stages, rather than by directly interpreting the internal representations or decision-making mechanisms of the underlying policy.
(b) It is also important to note that we do not establish or claim formal correctness of the generated spatiotemporal representations; instead, our objective is to empirically study whether the automatic derivation of interpretable, skill-level monitoring graphs from nominal visual execution trajectories using VLM- and LLM-based semantic extraction is a feasible approach.
(c) By analysing nominal execution episodes, our method extracts \emph{candidate} semantic constraints that characterise observed skill behaviour; as these constraints are inferred from a finite set of successful executions, they do not necessarily encompass all valid execution variations and consequently require validation before being treated as general execution requirements, for instance using causal discovery methods such as \cite{lee2023}.
(d) Finally, regarding the generalisation aspect, it should be noted that our work does not consider the question of \emph{when} it is actually feasible to generalise a model to another context; this question is beyond the scope of our study, but prior studies have shown that generalisation can be supported by incorporating semantic knowledge about scenario similarities, for instance an object ontology that can be used to compute similarities between objects as in \cite{mitrevski2023}.

%% file: subfiles/related-work.tex
\section{Related Work}
\label{sec:related-work}

As our objective is to use observations for extracting a semantic representation of the skill execution process that could subsequently be used for execution monitoring, our work has a close relation to work on learning semantic task representations, literature that uses temporal representations such as LTL to describe and verify execution, as well as work on execution monitoring.

\subsection{Semantic Robot Task Representations}

Ramirez-Amaro et al. \cite{ramirezamaro2017} learn tasks by observing human demonstrations, such that the main objective is to perform task segmentation and then map the extracted segments into primitive actions that the robot can reproduce.
Aein et al. \cite{aein2019} present a similar learning from demonstration approach where task segmentation is performed based on different object interaction stages, and reproduction is then performed by predefined primitives.
Zhang et al. \cite{zhang2026} discuss the use of symbolic knowledge in the context of hierarchical reinforcement learning, where symbolic fluents are used for learning high-level planning operators that are then associated with learned low-level policies that perform execution.

In this related work, the focus of semantic descriptions remains at the task level, such that the actual skill execution is presumed to be performed by well-defined motion primitives.
Our work is distinct from these methods as it focuses on semantically describing the lower-level skill execution process rather than focusing on the task level; in addition, our method does not require human observations or a dedicated reinforcement learning process, but works by processing data that is used for policy learning using imitation learning.

\subsection{Temporal Execution Representations}

Neider and Gavran \cite{neider2018} and Ghiorzi et al. \cite{ghiorzi2023} present algorithms which, given observations of fluents that represent positive and negative examples of a task of interest, can be used to extract an LTL formula that satisfies the positive examples and does not violate the negative examples.
Zudaire et al. \cite{zudaire2021} present a framework that makes it possible to specify conditions that should hold during a robot's execution process, and then uses those conditions for runtime verification.
The problem of automatic conversion of language commands to LTL formulas is addressed in Pan et al. \cite{pan2023}, where a language model is fine-tuned for this task.
In \cite{pek2023}, Pek et al. introduce SpaTiaL, a robot-agnostic framework for specifying and monitoring object-centric spatial and temporal constraints.
In \cite{quartey2025}, Quartey et al. propose a method that synthesises an executable robot plan based on a task description, with a focus on tasks that involve navigation and manipulation, such that plans are described by a generated LTL formula that is converted into an executable state machine.
A similar method is presented by Argenziano et al. \cite{argenziano2025}, where a task command is converted into an LTL form that is then used as a goal specification for creating a task plan, but also to support language-based queries about the execution process.
The method in Grigorev et al. \cite{grigorev2025} is also aligned with \cite{quartey2025,argenziano2025}, as language commands are converted into an LTL formula, but the LTL formula is used as a specification of constraints that are used for identifying problems with an already generated action sequence and suggesting corrections.
In \cite{wu2025}, Wu et al. describe a method that converts a task specification along with a set of task constraints into an LTL formula, and then the associated B{\"u}chi automaton is used to enforce constraint satisfaction during LLM-based plan generation.
Baert et al. \cite{baert2025} propose learning interpretable probabilistic automata that represent temporal task structures from demonstrations.
Guo et al. \cite{guo2025} propose a procedure based on which a PDDL planning problem is automatically generated from goals and Python scripts are generated for execution constraints, and both the problem and the constraint scripts are then passed on to a task and motion planner.
Finally, Stelter et al. \cite{stelter2026} introduce motion statecharts, which enables robot motions and monitors to be synthesised into an executable, state machine-like form.

In contrast to \cite{neider2018,ghiorzi2023}, our objective is to extract a temporal representation of a robot's execution process only from nominal observations, particularly because the variations of failures that could happen during robot execution is difficult to enumerate exhaustively.
As in \cite{zudaire2021}, our work also aims to obtain constraints that can be compiled into a monitorable representation; however, we extract these properties automatically by analysing execution traces rather than specifying them manually.
Unlike SpaTiaL \cite{pek2023}, our framework automatically extracts execution constraints from observational data rather than relying on explicitly provided specifications.
Unlike \cite{pan2023} and given that we are interested in the skill execution itself, we generate execution summaries (in natural language or as LTL sentences) by considering both task specifications and execution annotations.
Similar to \cite{quartey2025,argenziano2025,grigorev2025}, we aim to obtain an execution description that can be converted into a monitorable (graph) representation; however, our concrete focus is on the low-level skill execution process, and we also compare LTL and natural language summaries as execution description formats.
At a high level, \cite{wu2025} is aligned with our work; however, we address summary extraction and graph generation at a lower, skill-execution level.
As opposed to \cite{baert2025}, our focus is on representing intermediate skill execution stages through semantic relations inferred from visual observations.
As in \cite{guo2025}, our intention is to generate a monitorable representation from execution constraints, but we do not assume that a motion planner is used, but rather consider a more general case where execution may be performed using learned skills.
Unlike approaches that derive temporal or planning constraints primarily from natural-language task specifications \cite{wu2025,guo2025}, our framework infers semantic execution stages by analysing nominal, visual skill execution trajectories, without requiring those stages to be explicitly defined beforehand.
Conceptually, the approach in \cite{stelter2026} is also aligned with ours, as this is also an interpretable representation of the robot execution process, but we extract execution stages that define execution constraints automatically by analysing robot execution episodes rather than specifying those manually.

\subsection{Learning-Based Approaches for Execution Monitoring}

Liu et al. \cite{liu2023} describe a method that identifies failures and performs recovery using an LLM, such that the method first identifies whether subtasks are satisfied and, if they are not, attempts to find the reasons for the failure and performs recovery planning; to do this, the LLM is fed with a detailed scene graph description that summarises the robot's observations, along with a description of key frames during execution as well as the currently pursued subgoal.
Agia et al. \cite{agia2024} present a framework for monitoring learning-based policies, with a focus on detecting failures in the form of unstable behaviour and failures in making task progress; dedicated monitoring methods are defined for both types of failures.
In \cite{santhanam2026}, Santhanam et al. propose a method based on which observations collected during nominal executions are used to learn a model that is then used to detect online deviations, which then triggers a three-stage recovery process.
In \cite{sliwowski2025}, Sliwowski et al. describe ConditionNET, an execution monitoring approach that, given an observation and a task description, attempts to recognise the current execution stage in terms of the action's preconditions or effects; here, skills are represented as behaviour trees, and recovery behaviours are performed if the preconditions or effects are not satisfied.
Willibald and Lee \cite{willibald2026} present a human-in-the-loop task and motion learning method that performs skill segmentation based on inferred subgoals and feature clusters, and then uses subgoals to monitor the execution.
Thoduka et al. \cite{thoduka2025} introduce an execution failure detection model that is defined as an action classifier and which uses information about the sequence of actions during the execution to improve the classification.
In \cite{zhou2025}, Zhou et al. propose Code-as-Monitor, a framework that automatically extracts subgoals and associated execution constraints; for each of these constraints, monitoring code is generated and then embedded in the execution process for failure prediction.

Existing learning-based monitoring approaches differ in how execution constraints are represented and extracted.
ConditionNET \cite{sliwowski2025} learns visual representations of action preconditions and effects for execution-phase recognition and anomaly detection, Code-as-Monitor \cite{zhou2025} uses VLMs to generate executable spatiotemporal monitoring constraints from task information and visual observations, while Santhanam et al. \cite{santhanam2026} learn models of nominal execution behavior to detect anomalies and support recovery.
In contrast, our framework analyses nominal skill-execution trajectories to automatically extract semantic execution constraints through VLM-based annotations and LLM-based summarisation.
These constraints are subsequently converted into explicit monitoring graphs representing intermediate execution stages.
Our primary distinction is therefore the generation of interpretable, reusable skill-level monitoring representations, rather than directly learning anomaly detectors or generating executable monitoring code.

Similar to \cite{liu2023}, we also aim to benefit from the powerful reasoning capabilities of LLMs (and VLMs) for understanding a robot's behaviour; however, our concrete objective is to perform offline extraction of semantic information about the execution process that could then be used during execution monitoring.
As in \cite{agia2024}, our focus is also on obtaining a representation that can be used for recognising failures in task progress, but we aim to extract an explicit semantic representation of execution stages that can be used for this purpose, which has the ultimate objective of capturing richer, interpretable knowledge about the execution.
Similar to \cite{sliwowski2025}, our goal is to result in an execution representation that splits the process into different monitorable stages; however, unlike \cite{sliwowski2025}, which suspends anomaly detection during the core motion phase, our framework aims to monitor intermediate semantic states throughout skill execution, including the physical interaction phase.
In our work, instead of defining monitors in terms of goal regions that need to be reached as in \cite{willibald2026}, we define them in terms of automatically extracted execution states that are monitorable based on visual input.
In contrast to \cite{thoduka2025}, where the different execution stages are predefined, our objective is to automatically identify them; in addition, we aim to enable the use of semantic information about different execution stages for continuous monitoring rather than for post-execution classification.
Finally, \cite{zhou2025} requires richer information than ours, as it performs object segmentation for the purpose of subgoal identification and monitor generation; on the other hand, we investigate the possibility of extracting execution stages based on step-by-step execution annotations and without explicitly segmented objects of interest.

%% file: subfiles/methodology.tex
\section{Methodology}
\label{sec:methodology}

In this paper, we propose an execution monitoring system that automatically extracts monitors based on semantic summarisation of robot execution traces.
The objective of our work is to create a policy-agnostic monitoring framework that can be easily combined with off-the-shelf or custom-trained robot policies, as well as to improve the transparency of the execution process by mapping observations to concrete semantic stages.
We assume access to a dataset with nominal robot execution samples (this can be an imitation learning dataset or a dataset of nominal execution examples from an already learned policy), such that we use (i) a vision-language model (VLM) to automatically annotate individual steps of execution episodes and (ii) a large-language model (LLM) to extract a summary of the execution process given the annotations.
We then use an LLM to convert the summary into an execution graph that is finally embedded into the execution process as an external monitor.

\subsection{Formal Description}
\label{sec:formalisation}

Let $\skill$ be a set of robot skills (e.g. \texttt{pick-up}, \texttt{push}, and so forth), and $n$ be the number of skills of interest, namely we have $\skill_i \in \skill, 1 \leq i \leq n$, where $n\in\mathbb N_{>0}$.\footnote{From a formal point of view, we assume that a skill follows the definition in \cite{konidaris2018}, namely a skill $\skill$ is defined as $\skill = \left( \skillinit, \skillterm, \pi \right)$, where $\skillinit$ defines conditions under which the skill can be applied, $\skillterm$ defines conditions under which the skill is considered to have completed, and $\pi$ is an execution policy for the skill.}$^,$\footnote{Without loss of generality, we consider a fixed $n$ for the purposes of this work.}
Let $\mathcal{D}$ be a robot learning dataset containing imitation learning trajectories of the skills $\skill_i$, and $s_i\in\mathbb N_{>0}$ be the number of available trajectories of $\skill_i$ in $\mathcal{D}$.
Each trajectory $\mathcal{T}_i^j, 1 \leq j \leq s_i$ is given as a sequence of states $\vec{s} \in S$ and applied actions $\vec{a} \in A$ at each time step $t$:
\begin{equation}
    \mathcal{T}_i^j = \left[ \left( \vec{s}, \vec{a} \right)^{j}_{i,1}, \dots, \left( \vec{s}, \vec{a} \right)^{j}_{i,t_i^j} \right]
\end{equation}
where $t_i^j$ is the length of $\mathcal{T}_i^j$.
In this work, we consider the state $S$ to be represented by the space of images $I \subset \mathbb{R}^{H \times W \times 3}$, so the trajectories are given as
\begin{equation}
    \mathcal{T}_i^j = \left[ \left( I, \vec{a} \right)^{j}_{i,1}, \dots, \left( I, \vec{a} \right)^{j}_{i,t_i^j} \right]
\end{equation}
As is common in the literature on robot learning, we define the actions $\vec{a} \in \mathbb{R}^6$ as Cartesian end-effector motion commands.
It should, however, be noted that our proposed method analyses the skill execution process from observations, so it does not explicitly process the actions $\vec{a}$.

Our objective is twofold.
We first want to find a description $\mathcal{E}_{\skill_i}$ that defines semantic constraints on the execution process of $\skill_i$ based on the observations in $\mathcal{D}$ as well as any available prior knowledge.
Given $\mathcal{E}_{\skill_i}$, we then want to create a graph $\mathcal{G}_{\skill_i}$ of execution states that can then be used for online execution monitoring of $\skill_i$.

\subsection{Description Vocabulary}
\label{sec:description_vocabulary}

A description $\mathcal{E}_{\skill_i}$ that is fully open-ended may potentially refer to conditions that are either not possible to perceive by a robot or which are otherwise too general to be of use for execution monitoring.
To avoid this problem, we constrain the set of possible descriptions by using a \emph{description vocabulary} $V$, which includes terms that should be used in $\mathcal{E}_{\skill_i}$.
We define $V$ as a set of lifted predicates $r_k, 1 \leq k \leq \rho$, each of which may have a maximum arity of $q$:
\begin{equation}
    V = \{ r_k(X_1, \dots, X_q) \rightarrow \{ \top, \bot \} \; | \; 1 \leq k \leq \rho \}
\end{equation}
where $X_l \in \mathcal{X}, 1 \leq l \leq q$ are variables that should be grounded in $\mathcal{E}_{\skill_i}$, and $\mathcal{X}$ is the execution domain.
Examples of valid predicates are $visible(X)$ or $on(X_1, X_2)$.
This definition follows typical conventions in the literature on semantic representations for robotics \cite{hoefer2016,ramirezamaro2017,konidaris2018,mitrevski2023,diehl2023,zhang2026}.
Without loss of generality and in line with the related work, we assume that $q$ is maximum two, namely we only consider unary and binary predicates for the purposes of this work; thus, unary predicates are defined as $r_k:\mathcal X\rightarrow\{\top,\bot\}$ and binary predicates as $r_k:\mathcal X\times\mathcal X \rightarrow\{\top,\bot\}$.

\subsection{Execution Step Annotation}
\label{sec:step_description}

Our objective is to extract a semantic description of the robot's execution process of $\skill_i$, such that semantic information is extracted from the observable scene during execution.
Specifically, we consider a model $f$ that can provide a task-dependent semantic scene description $D^{j}_{i,t}$ given as
\begin{equation}
    D^{j}_{i,t} = f\left( I^{j}_{i,t}, L_i, V \right)
\end{equation}
where the description is provided for a skill specification $L_i$ and in terms of the predefined vocabulary $V$.
Concretely, let $\mathcal{L}=\Sigma^*$ denote the set of finite text strings over a token alphabet $\Sigma$, and let $\mathcal V$ denote the set of admissible predicate vocabularies. Then, the query function is defined as $f:\mathcal I\times\mathcal L\times \mathcal V\rightarrow\mathcal L$.

Given that $f$ needs to process both images $I$ as well as language-based specifications, a VLM is a natural choice for this task.
Considering the spatial nature of robotics tasks, we particularly use a model that is fine-tuned for spatial reasoning, similar to SpatialVLM \cite{chen2024}.

Given $\mathcal{T}_i^j$, we query $f$ for all steps $1 \leq t \leq t_i^j$ for which there is observable arm motion, namely steps where $\lVert \vec{a}^{j}_{i,t} \rVert > 0$.
We use the following \emph{step annotation prompt $P_D$} for this process:
\begin{mdframed}[style=roundframe]
    $P_D$ = \emph{A robot is trying to \textcolor{textred}{$L_i$}. Provide a description of what the robot is doing in the current scene. Provide the description in terms of relations that involve the robot. Use the following unary and binary relations that the robot knows about in your description: \textcolor{textred}{$V$}}
\end{mdframed}
Here, we constrain the annotations to only include relations involving the robot; without this constraint, the annotations would generally include scene relations that are inconsequential for the execution process or which are difficult to perceive in a general context.

This process results in a sequence $\mathcal{D}_{i}^j = \left[ D^{j}_{i,1}, \dots, D^{j}_{i,t^j_i} \right]\in\mathcal{L}^{t_i^j}$ of execution step descriptions.
$\mathcal{D}_{i}^j$ can be seen as a story that describes the complete execution based on the observations in $\mathcal{T}_i^j$.
Conceptually, this has similarities to the narrative-enabled episodic memories (NEEMs) in KnowRob 2.0 \cite{beetz2018}, but $\mathcal{D}_{i}^j$ is extracted purely from observations and targets the lower-level skill execution process rather than the higher-level task execution sequence that NEEMs describe.

\subsection{Episode Summary}
\label{sec:episode_summary}

The sequence $D^{j}_{i}$ is a step-by-step narration of the $j$-th execution episode of $\skill_i$, but is not, by itself, a description of the execution process that could be used for execution monitoring, as it contains information that is too fine-grained and consequently too specific for this purpose.
We thus consider a model $g$ that uses $D^{j}_{i}$ to extract a summary $\vec{m}^j_i$ given $L_i$ and in terms of $V$:
\begin{equation}
    \vec{m}^j_i = g\left( D^{j}_i, L_i, V \right)
\end{equation} 
As $g$ only processes language input, we use an LLM for summary extraction.
In this work, we investigate the generation of two types of summaries of the execution process: (i) a natural language summary and (ii) a summary in the form of a linear temporal logic (LTL) sentence \cite{pnueli1977}.

\paragraph{Natural language summary} The first type of summary $\vec{m}^j_i$ we consider is expressed in natural language.
We consider this both because LLMs are mostly trained on natural language, which means that it is likely that this is an easier summary to extract, but also because it could make it easier for VLMs to process the summary for monitoring.
To extract a language summary, we use the following \emph{language summary extraction prompt $\hat{P}_L$}:
\begin{mdframed}[style=roundframe]
    $\hat{P}_L = $ \emph{We have a list of observations of a robot performing the task \textcolor{textred}{$L_i$}. Use the observations to extract a one-sentence summary of conditions that should hold during the execution. The conditions should only refer to known relations from the following list: \textcolor{textred}{$V$}. The summary should be your only output. The observations are as follows: \textcolor{textred}{$D_i^j$}}
\end{mdframed}
This summary is still generated with respect to $V$ and, as we do not constrain the specific format, it is possible that $\vec{m}^j_i$ includes references to ground relations from $V$; such a case is, in fact, illustrated in Fig. \ref{fig:overview}.
Furthermore, we do not provide any grounding guidance in the summarisation prompt, as we want to investigate the models' inherent ability to perform correct grounding.

\paragraph{Summary expressed in LTL} The second type of summary $\vec{m}_j^i$ we explore is in the form of a ground LTL sentence; this is in line with recent work that uses LTL for representing temporal and monitorable robot plans \cite{argenziano2025,zudaire2021,ghiorzi2023,pan2023,quartey2025}.
We use the following \emph{LTL summary extraction prompt $\hat{P}_{LTL}$} for extracting an LTL sentence:
\begin{mdframed}[style=roundframe]
    $\hat{P}_{LTL} = $ \emph{We have a list of observations of a robot performing the task \textcolor{textred}{$L_i$}. Use the observations to extract a linear temporal logic formula that describes the execution. The formula should only include known relations from the following list: \textcolor{textred}{$V$}. The formula should be your only output. The observations are as follows: \textcolor{textred}{$D_i^j$}}
\end{mdframed}
For the same reason as $\hat{P}_L$, $\hat{P}_{LTL}$ does not include any grounding guidance or explicit constraints on the syntax of the LTL formula.

Considering these two summary representations, we can define $g$ in a general form as $g:\mathcal{L}^*\times\mathcal L \times\mathcal V \rightarrow \mathcal{M}$, where $\mathcal{L}^*=\bigcup_{m\geq0}\mathcal{L}^m$ denotes the set of finite sequences of text descriptions, $\mathcal{M}=\mathcal L\cup\Phi_{\mathrm{LTL}}$ denotes the possible summary representations, and $\Phi_{\mathrm{LTL}}$ is the set of syntactically valid LTL formulas over the permitted grounded predicates.

\subsection{Skill Summary: Semantic Envelope}
\label{sec:skill_summary}

Each $\vec{m}^j_i$ represents a summary of one specific execution episode of $\skill_i$; however, the particular way in which $\skill_i$ is executed may differ between episodes, namely $\mathcal{T}_i^j$ provides just one example of how the skill is executed, and no individual $\vec{m}^j_i$ is necessarily a good representation of the general execution process.
A representative summary should, instead, capture all $\vec{m}^j_i$ into a common form that can be applied to the skill as a whole.
We formalise this concept through the notion of a \emph{semantic envelope $\mathcal{E}_{\skill_i}$}:
\begin{mdframed}[style=roundblueframe]
    \begin{definition}
        Given a vocabulary $V$ as well as sentences $\vec{m}^j_i, 1 \leq j \leq s_i$ which describe the conditions that hold during the execution of $s_i$ episodes of skill $\skill_i$, a \emph{semantic envelope $\mathcal{E}_{\skill_i}$} is a specification of general constraints that should hold during the execution of $\skill_i$.
    \end{definition}
\end{mdframed}
Conceptually, $\mathcal{E}_{\skill_i}$ has some similarities to the trajectory envelope in \cite{haidu2015}; however, instead of the envelope being defined on the actions, we define $\mathcal{E}_{\skill_i}$ by conceptual criteria that should hold at different stages of the execution.

As every $\vec{m}^j_i$ is expressed in terms of language, we use the model $g$, which was already used for extracting $\vec{m}^j_i$, to extract $\mathcal{E}_{\skill_i}$ as well:
\begin{equation}
    \mathcal{E}_{\skill_i} = g\left( \vec{m}^{1:s_i}_i, L_i, V \right)
\end{equation}
where $\mathbf m_i^{1:s_i} = \left[\mathbf m_i^1,\ldots,\mathbf m_i^{s_i}\right]\in\mathcal M^{s_i}$ is the sequence of episode summaries.
Depending on whether the summaries $\vec{m}^j_i$ are expressed in natural language or as LTL formulas, we use the following \emph{envelope extraction prompts} $P_L$ and $P_{LTL}$ respectively:
\begin{mdframed}[style=roundframe]
    $P_L = $ \emph{A robot needs to perform the task \textcolor{textred}{$L_i$}. For \textcolor{textred}{$s_i$} execution episodes, we have the following list of summaries of conditions that held during each respective episode: \textcolor{textred}{$\vec{m}^{1:s_i}_i$}. Based on these summaries, extract a one-sentence summary of conditions that should hold during the execution of the skill. The conditions should only refer to known relations from the following list: \textcolor{textred}{$V$}. The summary should be your only output.}
\end{mdframed}

\begin{mdframed}[style=roundframe]
    $P_{LTL} = $ \emph{A robot needs to perform the task \textcolor{textred}{$L_i$}. For \textcolor{textred}{$s_i$} execution episodes, we have the following list of linear temporal logic formulas summarising the conditions that held during each respective episode: \textcolor{textred}{$\vec{m}^{1:s_i}_i$}. Based on these summaries, extract a linear temporal logic formula that describes the execution. The formula should only include known relations from the following list: \textcolor{textred}{$V$}. The formula should be your only output.}
\end{mdframed}

\subsection{Monitoring}
\label{sec:monitoring}

\begin{algorithm}[t]
    \begin{algorithmic}[1]
        \Function{\texttt{extractEnvelopes}}{$\mathcal{T}_i$, $L_i$, $V$, $P_D$, $\hat{P}_X$, $P_X$, $P_{\mathcal{G}}$, $P_{NER}$}
            \State $\vec{m}_i$ \assign $[]$
            \For{$j$ \algin $1$ \algto $s_i$}
                \State $D_i^j$ \assign $[]$
                \For{$t$ \algin $1$ \algto $t_i^j$}
                    \If{$\lVert \vec{a}^{j}_{i,t} \rVert > 0$}
                        \State $D_i^j$ \assign $D_i^j \; \mathbin{++} \; \left[f(I^j_{i,t}, L_i, V, P_D)\right]$
                    \EndIf
                \EndFor
                \State $\vec{m}_i$ \assign $\vec{m}_i \; \mathbin{++} \; \left[g( D^{j}_i, L_i, V, \hat{P}_X)\right]$
            \EndFor
            \State $\mathcal{E}_{\skill_i}$ \assign $g(\vec{m}_i, L_i, V, P_X)$
            \State $\mathcal{G}_{\skill_i}$ \assign $g(\mathcal{E}_{\skill_i}, L_i)$
            \State $(X_{1:\gamma}, entities_{1:\gamma})$ \assign $g(L_i, P_{NER})$
            \State $\mathcal{E}_{\skill_i}$ \assign $\operatorname{lift}(\mathcal{E}_{\skill_i}, entities_{1:\gamma}, X_{1:\gamma})$
            \State $\mathcal{G}_{\skill_i}$ \assign $\operatorname{lift}(\mathcal{G}_{\skill_i}, entities_{1:\gamma}, X_{1:\gamma})$
            \State \Return ($\mathcal{E}_{\skill_i}$, $\mathcal{G}_{\skill_i}$)
        \EndFunction
    \end{algorithmic}
    \caption{A summary of our method for extracting semantic envelopes and monitoring graphs by processing nominal execution episodes. The function $f$ is represented by a VLM, while the function $g$ by an LLM (both thus take the prompt as an additional explicit argument). $\hat{P}_X$ and $P_X$ are set as described in the text, depending on whether the envelope is represented by a language-based summary or by an LTL sentence. The operator $\mathbin{++}$ concatenates two lists.}
    \label{alg:monitor_extraction_alg}
\end{algorithm}

The envelope $\mathcal{E}_{\skill_i}$is not, by itself, a suitable monitoring representation, as it does not decompose the execution into monitorable execution stages.
To make it both suitable for execution monitoring and interpretable, we convert $\mathcal{E}_{\skill_i}$ into a \emph{labelled graph} representation
\begin{equation}
    \mathcal{G}_{\skill_i} = (N_{\skill_i}, E_{\skill_i}, R_{\skill_i})
\end{equation}
where the nodes $N_{\skill_i}$ represent different execution stages, the edges $E_{\skill_i}: N_{\skill_i} \rightarrow N_{\skill_i}$ are possible transitions, and the node annotations $R_{\skill_i}: \left\{ r_{z}(x_1, \dots, x_q) \rightarrow \{ \top, \bot \} \; | z \in \mathbb{N} \; \right\}$ denote ground relations that hold in each respective stage.\footnote{This definition of $\mathcal{G}_{\skill_i}$ is inspired by a B{\"u}chi automaton \cite{buechi1966}; however, we do not intend to claim formal correctness of the graph as necessary for LTL model checking given that we use LLMs for graph extraction, so we define $\mathcal{G}_{\skill_i}$ as a general labelled transition system instead.}

Keeping in line with the procedure for extracting $\vec{m}^j_i$ and $\mathcal{E}_{\skill_i}$, we use $g$ to convert $\mathcal{E}_{\skill_i}$ into $\mathcal{G}_{\skill_i}$.
We do this using the following \emph{envelope-to-graph conversion prompt $P_{\mathcal{G}}$}:
\begin{mdframed}[style=roundframe]
    $P_{\mathcal{G}}$ = \emph{Convert the following temporal description of a task \textcolor{textred}{$L_i$} into graph states to describe the conditions that need to hold in each execution state: \textcolor{textred}{$\mathcal{E}_{\skill_i}$}. Assign meaningful names to all states. Your output should be in the form of a Python dictionary. The dictionary should also include an entry `transitions' that includes valid transitions between the states. The dictionary should be your only output. Remove all display formatting in the output.}
\end{mdframed}

Given $\mathcal{G}_{\skill_i}$, the final aspect of our framework is concerned with execution monitoring.
We consider a historical horizon $h$ of observations, namely $I_{t-h:t}$, and use a VLM $f_{M}$ as a monitor, where $f_M$ is not necessarily the same as $f$.\footnote{As described in the following section, we use two different models for annotation and monitoring due to an observed difference in performance.}$^{,}$\footnote{It should be noted that, for the purposes of this work, we only aim to demonstrate the feasibility of using $\mathcal{G}_{\skill_i}$ for monitoring rather than ensuring that the monitoring algorithm can be used in real time; thus, in the evaluation in the following section, we do not consider the monitoring frequency as an evaluation factor.}
Similar to \cite{sliwowski2025}, the monitor's objective is recognising the execution stage at time $t$, expressed in terms of $N_{\skill_i}$ along with an additional $\mathrm{uncertain}$ state that may indicate $f_M$'s inability to definitively recognise the current execution stage $N_{\mathcal S_i,t}$:
\begin{equation}
    N_{\skill_i,t}= f_M(I_{t-h:t}, L_i, \mathcal{G}_{\skill_i})
\end{equation}
where $f_M: \mathcal{I}^{h+1} \times \mathcal{L} \times \mathbb{G} \rightarrow N_{\mathcal S_i} \cup \{\mathrm{uncertain}\}$, with $\mathbb{G}$ as the set of admissible monitoring graphs.
The $\mathrm{uncertain}$ state is included in $N_{\skill_i}$ with \emph{no valid incoming transitions from the other nodes} (a transition to this state could thus be used as an indication of a possible failure) and \emph{valid outgoing transitions to all other nodes} (this allows $f_M$ to recover from any actual execution state).

For monitoring, the inputs to $f_M$ are $I_{t-h:t}$ (represented as one image with the frames arranged from left to right) along with a \emph{monitoring prompt $P_M$}:
\begin{mdframed}[style=roundframe]
    $P_M$ = \emph{We have a sequence of images of a robot executing an ongoing task \textcolor{textred}{$L_i$}. The rightmost image is the current observation, while the other images are the previous observations. The following is a description of states that may hold during or after the execution: \textcolor{textred}{$\mathcal{G}_{\skill_i}$}. Which of the states is the execution in? You have to consider all images in your answer. If the scene is not clearly visible, answer with 'uncertain'; don't assume the answer. The name of the state should be your only output.}
\end{mdframed}
Given $N_{\skill_i,t}$, the monitor can be used to trigger an alarm, which we define as an event triggered by an unexpected transition from $N_{\skill_i,t-1}$ to $N_{\skill_i,t}$.
Formally, this simplified alarm rule is defined as:
\begin{equation}
    a_t =
    \begin{cases}
        1, &(N_{\skill_i,t-1}, N_{\skill_i,t})\notin E_{\mathcal S_i}, \\
        0, &(N_{\skill_i,t-1}, N_{\skill_i,t})\in E_{\mathcal S_i}
    \end{cases}
\end{equation}
As there are no valid transitions to the $\mathrm{uncertain}$ state in $\mathcal{G}_{\skill_i}$, this rule treats observed transitions to it as alarms.\footnote{In practice, it is better to perform filtering on the observed state transitions rather than raise an alarm immediately, as otherwise noisy estimates of the execution stage would unnecessarily affect the execution process (we illustrate such false alarms in Sec. \ref{sec:evaluation:monitoring}).}

Here, it should be reiterated that the generated monitoring graph provides an interpretable abstraction of the execution stages inferred from the semantic envelope; however, since the graph generation relies on LLM inference, the resulting state and transition structure is not guaranteed to be semantically equivalent to the original language-based or LTL envelope.
Consequently, our proposed monitoring procedure identifies deviations from the generated execution stage transitions rather than verifying satisfaction of the complete temporal specification.

\subsection{Envelope and Monitor Generalisation}
\label{sec:envelope_generalisation}

The $\mathcal{E}_{\skill_i}$ and $\mathcal{G}_{\skill_i}$ extracted above are grounded; this is because $\mathcal{D}_{i}^j$ and subsequently also $\vec{m}^{1:s_i}_i$ refer to objects that are observed in the execution scene.
This grounding makes the envelope and the monitoring graph specific to the execution case that is observed $\mathcal{T}_i^j$; however, ideally, we would like the envelope to be generalisable across different objects.
To perform this generalisation, we create a lifted version of $\mathcal{E}_{\skill_i}$ and $\mathcal{G}_{\skill_i}$ in which the names of concrete entities are replaced by variables that need to be grounded at runtime.
We achieve this by
(i) performing named entity recognition \cite{jehangir2023} on $L_i$, during which named entities are associated with variable names $X_{1:\gamma}, \gamma \in \mathbb{N}$, and
(ii) substituting all instances of the entities in $\mathcal{E}_{\skill_i}$ and $\mathcal{G}_{\skill_i}$ with the variables $X_{1:\gamma}$.
For consistency with the rest of our framework, we perform named-entity recognition using an LLM, such that we use the following \emph{entity recognition prompt} $P_{NER}$ for this purpose:
\begin{mdframed}[style=roundframe]
    $P_{NER} = $ = \emph{Extract named entities from the following command: "\textcolor{textred}{$L_i$}". Your output should be in the form of a dictionary \{"X1": entity\_1, ..., "Xn": entity\_n\}..., where X1, ..., Xn are variables, and entity\_1, ..., entity\_n are grounded to real-world objects named in the sentence. Leave the variables unchanged in the output dictionary. The entities should only name objects. The dictionary should be your only output.}
\end{mdframed}
During online monitoring, the process is reversed, namely
(i) we recognise entities in $L_i$ and
(ii) ground $\mathcal{G}_{\skill_i}$ by applying the substitution $\{ X_1 / entity_1, ..., X_\gamma / entity_\gamma \}$.

Our method for extracting $\mathcal{E}_{\skill_i}$ and monitoring using $\mathcal{G}_{\skill_i}$ is summarised in Alg. \ref{alg:monitor_extraction_alg} and Alg. \ref{alg:monitoring} respectively.

\begin{algorithm}[t]
    \begin{algorithmic}[1]
        \Function{\texttt{monitorExecution}}{$L_i$, $\mathcal{G}_{\skill_i}$, $h$, $P_M$, $P_{NER}$}
            \State $\vec{n}$ \assign $[]$
            \State $alarms$ \assign $[]$
            \State $(X_{1:\gamma}, entities_{1:\gamma})$ \assign $g(L_i, P_{NER})$
            \State $\mathcal{G}_{\skill_i}$ \assign $\operatorname{ground}(\mathcal{G}_{\skill_i}, X_{1:\gamma}, entities_{1:\gamma})$
            \For{$t$ \algin $1$ \algto $\infty$}
                \State $\vec{n}$ \assign $\vec{n} \; \mathbin{++} \; [f_M(I_{t-h:t}, L_i, \mathcal{G}_{\skill_i}, P_M)]$
                \If{$(\vec{n}_{t-1}, \vec{n}_{t})$ \algnot \algin $E_{\skill_i}$}
                    \State $alarms$ \assign $alarms \; \mathbin{++} [\top]$
                \Else
                    \State $alarms$ \assign $alarms \; \mathbin{++} [\bot]$
                \EndIf
            \EndFor
        \EndFunction
    \end{algorithmic}
    \caption{A summary of our monitoring method given a graph extracted from the semantic envelope and a predefined monitoring history horizon.}
    \label{alg:monitoring}
\end{algorithm}

%% file: subfiles/evaluation.tex
\section{Evaluation}
\label{sec:evaluation}

\subsection{Evaluation Setup}

\begin{figure}[t]
    \begin{subfigure}[t]{0.195\linewidth}
        \includegraphics[width=\linewidth]{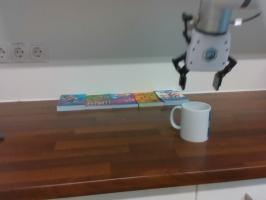}
        \caption{Picking a cup from a surface}
    \end{subfigure}
    \begin{subfigure}[t]{0.195\linewidth}
        \includegraphics[width=\linewidth]{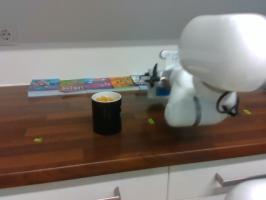}
        \caption{Placing a cup on a surface (next to another cup)}
    \end{subfigure}
    \begin{subfigure}[t]{0.195\linewidth}
        \includegraphics[width=\linewidth]{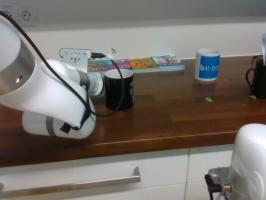}
        \caption{Pushing a cup}
    \end{subfigure}
    \begin{subfigure}[t]{0.195\linewidth}
        \includegraphics[width=\linewidth]{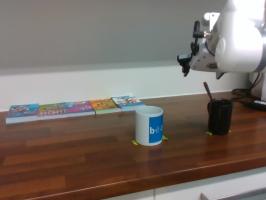}
        \caption{Picking a spoon from a cup}
    \end{subfigure}
    \begin{subfigure}[t]{0.195\linewidth}
        \includegraphics[width=\linewidth]{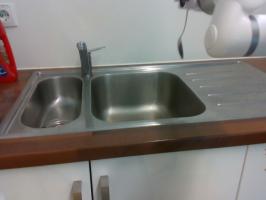}
        \caption{Placing a spoon inside a sink}
    \end{subfigure}
    \caption{Illustration of the five single-arm skills from the COFFAIL dataset \cite{coffail2026} which we use throughout the evaluation.}
    \label{fig:coffail-illustration}
\end{figure}

\paragraph{Evaluation data:}
We evaluate our proposed framework and the four research questions from Sec. \ref{sec:introduction} through a study focused on various everyday skills performed by a robot manipulator.
Throughout the analysis, we use the COFFAIL dataset \cite{coffail2026}, which includes data from seven common domestic skills collected with a bimanual manipulator (of these, five are single-arm skills and two are bimanual skills).
We use the data from the single-arm skills in our evaluation\footnote{We do not use the dual-arm skills in this paper primarily because our generalisation study uses a single-arm robot.} (picking a cup from a surface, placing a cup on a surface, pushing a cup, picking a spoon from a cup, and placing a spoon in a sink) and the observations from the robot's head camera.\footnote{The dataset also contains observations from a wrist camera, but we do not use these due to their limited view of the scene, which is less suitable for extracting semantic envelopes and subsequent monitoring.}
Fig. \ref{fig:coffail-illustration} illustrates these skills through images from the head camera view.

This dataset is particularly suitable for our study because it includes both nominal and some anomalous executions; this enables us to analyse our framework fully, namely from generating monitors from nominal data to monitoring on anomalous data.
While our method is not constrained to specific types of anomalies, the concrete anomalies we consider in the analysis include failed grasps and camera occlusions.

To evaluate the generalisation of the generated monitoring graphs, we use separate data collected with a PAL TIAGo robot\footnote{\url{https://pal-robotics.com/robot/tiago/}}, also from a head camera view.

\paragraph{Baselines:}
To quantify the advantage of our proposed method with respect to simple model prompting strategies that one might intuitively consider instead, we generate baseline envelopes by querying (i) LLMs without evidence from execution episodes, thereby essentially extracting implicit ontological knowledge about robot skills from the models, and (ii) VLMs with minimal evidence provided in terms of key execution frames.

\paragraph{Execution step annotation and semantic envelope extraction:}
In the evaluation, we use \emph{SpaceOm}\footnote{\url{https://huggingface.co/remyxai/SpaceOm}} as the model $f$ that annotates skill execution episodes; this model is based on SpatialVLM \cite{chen2024} and is thus specialised for spatial reasoning tasks, which we consider an essential property for robot skill execution.\footnote{We run this model locally on an Apple M1 Pro 16-Core GPU.}
The model $g$ used for extracting execution summaries $\vec{m}_i^j$ and semantic envelopes $\mathcal{E}_{\skill_i}$ is an LLM, such that, for comparison, we use multiple models in our evaluation: \emph{Claude Opus 4.6}, \emph{DeepSeek V4 Flash}, and \emph{Gemma 4 31B}.\footnote{We use the official Claude API for interacting with Claude.}$^{,}$\footnote{We use the Ollama API to interact with the cloud versions of DeepSeek and Gemma: \url{https://ollama.com/}.}

\paragraph{Evaluating summary correctness:}
To analyse the correctness of the generated summaries $\vec{m}^j_i$, we manually craft ground-truth execution summaries $\vec{g}_i$ for the skills of interest; these are given in Tab. \ref{tab:ground-truth-summaries}, along with the skill specifications $L_i$.
\begin{table}[t]
    \centering
    \caption{Skill specifications $L_i$ and ground-truth descriptions of the execution conditions for the skills used in the evaluation}
    \label{tab:ground-truth-summaries}
    \begin{tabular}{M{0.15\linewidth} | M{0.25\linewidth} | M{0.57\linewidth}}
        \cellcolor{blue!10}\textbf{Skill} & \cellcolor{blue!10}\textbf{$L_i$} & \cellcolor{blue!10}\textbf{Ground-truth summary} \\\hline
        \cellcolor{blue!10}Cup pickup     & \emph{Pick the cup} & \emph{The robot should be in front of the cup and move towards it, and should eventually hold the cup.} \\\hline
        \cellcolor{blue!10}Cup placing    & \emph{Place the cup you are holding} & \emph{The robot should hold the cup and be above the placing surface, moving towards the surface, and should eventually release the cup.} \\\hline
        \cellcolor{blue!10}Cup moving     & \emph{Move the cup} & \emph{The robot should hold and push the cup, and should eventually release the cup.} \\\hline
        \cellcolor{blue!10}Spoon picking from a cup  & \emph{Pick up the spoon} & \emph{The robot should be above and in front of the cup with the spoon, moving towards it, and should eventually hold the spoon.} \\\hline
        \cellcolor{blue!10}Spoon placing in a sink  & \emph{Place the spoon you are holding in the sink} & \emph{The robot should hold the spoon and be above the sink, moving towards the sink, and should eventually release the spoon.}
    \end{tabular}
\end{table}
Throughout the analysis, we use the following set of relations as the nominal description vocabulary $V$:
\begin{mdframed}[style=roundframe]
    \vspace{-0.4cm}
    \begin{equation*}
        \begin{split}
            V = \{ &\texttt{leftOf}(x, y),    \texttt{rightOf}(x, y),   \texttt{above}(x, y),    \texttt{below}(x, y), \texttt{inFrontOf}(x, y), \\
                   &\texttt{behind}(x, y),    \texttt{nextTo}(x, y),    \texttt{on}(x, y),       \texttt{in}(x, y),    \texttt{holding}(x, y), \\
                   & \texttt{pushing}(x, y),  \texttt{pulling}(x, y),   \texttt{cutting}(x, y),  \texttt{twisting}(x, y), \\
                   &\texttt{throwing}(x, y), \texttt{stirring}(x, y) \}
        \end{split}
    \end{equation*}
\end{mdframed}
This vocabulary includes both predicates that are relevant for describing the skills of interest as well as some irrelevant predicates, and thus makes it possible to study the models' sensitivity to unimportant information when generating annotations and execution summaries.

The generated natural-language based summaries $\vec{m}_i^j$ and language-based envelopes $\mathcal{E}_{\skill_i}$ can be directly compared with the ground-truth summaries in Tab. \ref{tab:ground-truth-summaries}.
To make it possible to compare the generated LTL summaries with the ground-truth descriptions as well, we convert the LTL sentences to a natural language summary.\footnote{This translation may potentially lead to loss of aspects that are actually represented by the formula, but should still indicate a general trend in the summary correctness: a very incorrect formula is likely to be translated into a language sentence that significantly differs from the ground truth.}
We do this by querying $g$ as follows:
\begin{mdframed}[style=roundframe]
    \emph{A robot is trying to perform the task \textcolor{textred}{$L_i$}. We are given the following LTL formula that describes the execution: \textcolor{textred}{LTL sentence}. Translate the formula into a one-sentence summary. The summary should be your only output.}
\end{mdframed}

To compare the generated and ground-truth descriptions, we map both descriptions to a common latent space with a sentence transformer \cite{sentencebert2019} $st$ and then use the cosine similarity to compare the vectors in the latent space\footnote{We concretely use the \emph{all-mpnet-base-v2} transformer, which has been benchmarked as a performant model across different tasks: \url{https://sbert.net/docs/sentence_transformer/pretrained_models.html}}:
\begin{equation}
    \operatorname{sim}(\vec{g}_i, \vec{m}^j_i) = \cos(\theta) = \frac{st\left( \vec{g}_i \right) \cdot st\left( \vec{m}^j_i \right)}{\lVert st\left( \vec{g}_i \right) \rVert \lVert st\left( \vec{m}^j_i \right) \rVert}
\end{equation}

\subsection{Baseline Summaries}

\begin{figure}[t]
    \begin{subfigure}[b]{0.475\linewidth}
        \includegraphics[width=\linewidth]{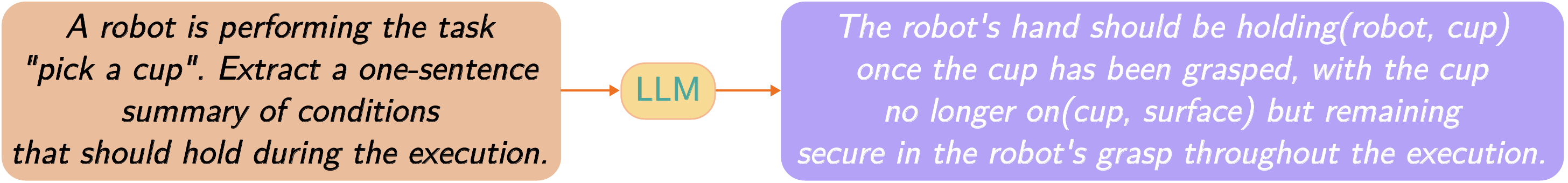}
        \vspace{0.35cm}
        \caption{Textual baseline}
    \end{subfigure}
    \hspace{0.025\linewidth}
    \begin{subfigure}[b]{0.475\linewidth}
        \includegraphics[width=\linewidth]{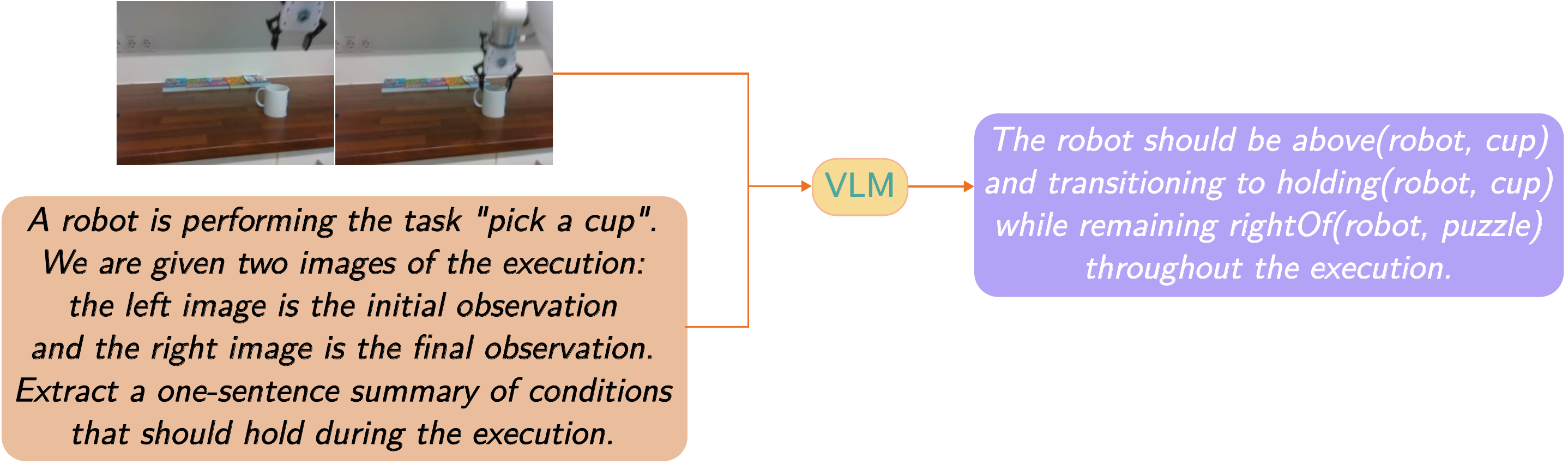}
        \caption{Visual baseline}
    \end{subfigure}
    \caption{Illustration of the two baselines for generating execution summaries, with examples of generated summaries in the cup pickup scenario.}
    \label{fig:baseline-illustration}
\end{figure}

We assess RQ1 (the feasibility of extracting semantic execution descriptions with VLMs and LLMs) by generating two types of baseline execution summaries:
(i) a \emph{textual baseline} for which we only provide the task description to an LLM, without any observations from concrete execution episodes, and
(ii) a \emph{visual baseline}, which we extract from a VLM by providing the initial and final observations of each episode (passed to the model as one image with the frames arranged next to each other) along with the task description.
The baselines are illustrated in Fig. \ref{fig:baseline-illustration}.
In the subsequent analysis, we use these summaries as a baseline for studying the effect of incorporating concrete observations in the summary generation.

We use the following prompt for extracting the textual baseline summaries (with options depending on the summary type):
\begin{mdframed}[style=roundframe]
    \emph{A robot is performing the task \textcolor{textred}{$L_i$}. Extract [a one-sentence summary of conditions that should hold during the execution] / [a linear temporal logic formula that describes the execution]. The [conditions should only refer to] / [formula should only include] known relations from the following list: \textcolor{textred}{$V$}. The [summary] / [formula] should be your only output.}
\end{mdframed}
To account for variability in outputs between calls, we generate each textual baseline summary $10$ times.

The visual baseline summaries are generated with the following prompt:
\begin{mdframed}[style=roundframe]
    \emph{A robot is performing the task \textcolor{textred}{$L_i$}. We are given two images of the execution: the left image is the initial observation and the right image is the final observation. Extract [a one-sentence summary of conditions that should hold during the execution] / [a linear temporal logic formula that describes the execution]. The [conditions should only refer to] / [formula should only include] known relations from the following list: \textcolor{textred}{$V$}. The [summary] / [formula] should be your only output.}
\end{mdframed}
We generate the visual baseline summaries for each nominal episode in the dataset.
It should be noted that the visual baseline requires a VLM, so we only use \emph{Claude Opus 4.6} and \emph{Gemma 4 31B} here; \emph{DeepSeek V4 Flash} is an LLM, so it cannot process visual input.

Tab. \ref{tab:baseline-similarities} shows the similarities between the ground-truth summaries and the textual baseline summaries, averaged over $10$ generated textual baseline summaries.
\begin{table}[t]
    \centering
    \scriptsize
    \caption{Textual baseline semantic similarities with the ground-truth summaries for different summary types}
    \label{tab:baseline-similarities}
    \begin{tabular}{M{0.12\linewidth} | M{0.135\linewidth} | M{0.135\linewidth} | M{0.135\linewidth} | M{0.135\linewidth} | M{0.135\linewidth} | M{0.135\linewidth}}
        \cellcolor{blue!10} & \multicolumn{2}{c}{\cellcolor{blue!10}\textbf{Claude Opus 4.6}} & \multicolumn{2}{c}{\cellcolor{blue!10}\textbf{DeepSeek V4 Flash}} & \multicolumn{2}{c}{\cellcolor{blue!10}\textbf{Gemma 4 31B}} \\\cline{2-7}
        \multirow{-2}{*}{\cellcolor{blue!10}\textbf{Skill}} & \cellcolor{blue!10}\textbf{Language summary} & \cellcolor{blue!10}\textbf{LTL formula} & \cellcolor{blue!10}\textbf{Language summary} & \cellcolor{blue!10}\textbf{LTL formula} & \cellcolor{blue!10}\textbf{Language summary} & \cellcolor{blue!10}\textbf{LTL formula} \\\hline
        \cellcolor{blue!10}Cup pickup     & $0.77 \pm 0.06$ & $0.86 \pm 0.01$ & $0.85 \pm 0.06$ & $0.88 \pm 0.01$ & $0.86 \pm 0.02$ & $0.88 \pm 0.01$ \\\hline
        \cellcolor{blue!10}Cup placing    & $0.92 \pm 0.01$ & $0.92 \pm 0.01$ & $0.78 \pm 0.05$ & $0.84 \pm 0.02$ & $0.84 \pm 0.05$ & $0.85 \pm 0.05$ \\\hline
        \cellcolor{blue!10}Cup moving     & $0.83 \pm 0.01$ & $0.91 \pm 0.02$ & $0.77 \pm 0.08$ & $0.85 \pm 0.13$ & $0.83 \pm 0.04$ & $0.90 \pm 0.02$ \\\hline
        \cellcolor{blue!10}Spoon picking  & $0.81 \pm 0.03$ & $0.82 \pm 0.01$ & $0.81 \pm 0.07$ & $0.81 \pm 0.01$ & $0.86 \pm 0.04$ & $0.82 \pm 0.01$ \\\hline
        \cellcolor{blue!10}Spoon placing  & $0.84 \pm 0.04$ & $0.91 \pm 0.02$ & $0.76 \pm 0.08$ & $0.86 \pm 0.01$ & $0.89 \pm 0.02$ & $0.87 \pm 0.05$ \\\hline
    \end{tabular}
\end{table}
The similarities between the ground-truth summaries and the visual baseline summaries are shown in Tab. \ref{tab:baseline-vision-similarities}, averaged over all skill episodes.
\begin{table}
    \centering
    \scriptsize
    \caption{Visual baseline semantic similarities with the ground-truth summaries for different summary types}
    \label{tab:baseline-vision-similarities}
    \begin{tabular}{M{0.12\linewidth} | M{0.21\linewidth} | M{0.21\linewidth} | M{0.21\linewidth} | M{0.21\linewidth}}
        \cellcolor{blue!10} & \multicolumn{2}{c}{\cellcolor{blue!10}\textbf{Claude Opus 4.6}} & \multicolumn{2}{c}{\cellcolor{blue!10}\textbf{Gemma 4 31B}} \\\cline{2-5}
        \multirow{-2}{*}{\cellcolor{blue!10}\textbf{Skill}} & \cellcolor{blue!10}\textbf{Language summary} & \cellcolor{blue!10}\textbf{LTL formula} & \cellcolor{blue!10}\textbf{Language summary} & \cellcolor{blue!10}\textbf{LTL formula} \\\hline
        \cellcolor{blue!10}Cup pickup     & $0.79 \pm 0.05$ & $0.83 \pm 0.03$ & $0.79 \pm 0.06$ & $0.88 \pm 0.02$ \\\hline
        \cellcolor{blue!10}Cup placing    & $0.82 \pm 0.05$ & $0.87 \pm 0.01$ & $0.84 \pm 0.02$ & $0.85 \pm 0.03$ \\\hline
        \cellcolor{blue!10}Cup moving     & $0.78 \pm 0.03$ & $0.87 \pm 0.02$ & $0.74 \pm 0.00$ & $0.85 \pm 0.04$ \\\hline
        \cellcolor{blue!10}Spoon picking  & $0.83 \pm 0.04$ & $0.80 \pm 0.01$ & $0.78 \pm 0.07$ & $0.84 \pm 0.02$ \\\hline
        \cellcolor{blue!10}Spoon placing  & $0.91 \pm 0.03$ & $0.91 \pm 0.01$ & $0.88 \pm 0.03$ & $0.90 \pm 0.01$ \\\hline
    \end{tabular}
\end{table}
\emph{Considering that the similarities with the ground-truth summaries for both baselines are, in the most cases, around or larger than $0.8$, the results confirm RQ1}; this is not an unexpected result, but it nevertheless demonstrates that the models have an ability to extract generally meaningful summaries even with limited information, although there is an observable variation between models and tasks.
For instance, as indicated by the results for the spoon picking skill, the correctness expectedly reduces when a skill is less generic and may require more concrete world knowledge to be executed correctly.

\subsection{Ability to Incorporate Evidence}

To investigate the models' ability to use episode step annotations when generating execution summaries, we use $V$ to annotate the execution steps in all nominal trajectories, and then generate the summaries $\vec{m}^j_i$ for all episodes.
As above, we compare the generated summaries with the ground-truth summaries; however, to study RQ2 (the benefit of including semantic evidence for summary generation), we additionally perform Welch's t-test \cite{welch1947} between the baseline (prior) similarities and the posterior similarities after the inclusion of observations, as this allows us to validate whether the inclusion of concrete execution evidence does indeed significantly change the similarities with the ground truth.
We perform this comparison between the prior and posterior similarities at two significance levels ($\alpha=0.05$ and $\alpha=0.01$) in order to see how the ability to incorporate evidence varies between skills and models.

Tab. \ref{tab:summary-type-evaluation} shows the similarities of the generated summaries with the ground-truth summaries, averaged over all $s_i$ trajectories for skill $\skill_i$, along with the results from the t-test.
\begin{table}[t]
    \centering
    \scriptsize
    \caption{Average semantic similarities with the ground-truth summaries (over all skill episodes) for different summary types and models. A $*$ denotes a statistically significant difference from the textual baseline at $\alpha=0.05$, and a $**$ a statistically significant difference at $\alpha=0.01$; similarly, an $\times$ denotes a statistically significant difference from the visual baseline at $\alpha=0.05$, and a $\times\times$ a statistically significant difference at $\alpha=0.01$.}
    \label{tab:summary-type-evaluation}
    \begin{tabular}{M{0.12\linewidth} | M{0.135\linewidth} | M{0.135\linewidth} | M{0.135\linewidth} | M{0.135\linewidth} | M{0.135\linewidth} | M{0.135\linewidth}}
        \cellcolor{blue!10} & \multicolumn{2}{c}{\cellcolor{blue!10}\textbf{Claude Opus 4.6}} & \multicolumn{2}{c}{\cellcolor{blue!10}\textbf{DeepSeek V4 Flash}} & \multicolumn{2}{c}{\cellcolor{blue!10}\textbf{Gemma 4 31B}} \\\cline{2-7}
        \multirow{-2}{*}{\cellcolor{blue!10}\textbf{Skill}} & \cellcolor{blue!10}\textbf{Language summary} & \cellcolor{blue!10}\textbf{LTL formula} & \cellcolor{blue!10}\textbf{Language summary} & \cellcolor{blue!10}\textbf{LTL formula} & \cellcolor{blue!10}\textbf{Language summary} & \cellcolor{blue!10}\textbf{LTL formula} \\\hline
        \cellcolor{blue!10}Cup pickup     & $0.86 \pm 0.02^{**}_{\times\times}$ & $0.94 \pm 0.01^{**}_{\times\times}$ & $0.88 \pm 0.06$ & $0.87 \pm 0.06$ & $0.86 \pm 0.04_{\times\times}$ & $0.97 \pm 0.00^{**}_{\times\times}$ \\\hline
        \cellcolor{blue!10}Cup placing    & $0.88 \pm 0.01^{**}_{\times\times}$ & $0.89 \pm 0.02^{**}_{\times}$ & $0.84 \pm 0.03^{**}$ & $0.85 \pm 0.02$ & $0.88 \pm 0.03^{*}_{\times\times}$ & $0.91 \pm 0.03^{*}_{\times\times}$ \\\hline
        \cellcolor{blue!10}Cup moving     & $0.86 \pm 0.01^{**}_{\times\times}$ & $0.85 \pm 0.05^{**}$ & $0.88 \pm 0.01^{**}$ & $0.82 \pm 0.10$ & $0.81 \pm 0.04_{\times\times}$ & $0.93 \pm 0.01^{**}_{\times\times}$ \\\hline
        \cellcolor{blue!10}Spoon picking  & $0.85 \pm 0.02^{*}$ & $0.91 \pm 0.02^{**}_{\times\times}$ & $0.87 \pm 0.04^{*}$ & $0.88 \pm 0.08^{*}$ & $0.85 \pm 0.05_{\times}$ & $0.91 \pm 0.02^{**}_{\times\times}$ \\\hline
        \cellcolor{blue!10}Spoon placing  & $0.91 \pm 0.02^{**}$ & $0.94 \pm 0.02^{*}_{\times\times}$ & $0.84 \pm 0.06^{*}$ & $0.91 \pm 0.03^{**}$ & $0.83 \pm 0.05^{**}_{\times\times}$ & $0.94 \pm 0.02^{**}_{\times\times}$ \\\hline
    \end{tabular}
\end{table}
As can be seen here, the evidence from the annotated observations $D_i^j$ leads to changes in the similarity with the ground truth that are statistically significant (at $\alpha = 0.01$) in the majority of cases (across summary types, though not equally noticeable across models).
\emph{This confirms RQ2, as it shows that the incorporation of detailed execution evidence as proposed in our work is beneficial to consider for generating skill execution summaries.}

It should, however, be noted that, in a few cases, the similarity with the ground truth reduces after the evidence incorporation; this is, for instance, the case for \emph{Claude Opus 4.6} in the cup placing case, where the similarity of the language-based summary reduces from $0.92$ with the textual baseline to $0.88$ after evidence incorporation.
Qualitatively, we could notice that this reduction is caused by differences in how objects in the environment are referred to; for instance, in the cup placing example, we generically refer to a \emph{placing surface} in our ground-truth description, which is also what the baseline summary refers to, but \emph{Claude Opus 4.6} refers (correctly based on the scene) to a placing counter.
This suggests that these reductions of similarity are referential rather than necessarily conceptual, at least for the skills of interest.\footnote{This problem could potentially be avoided by constraining the feasible object references, for instance based on ontological knowledge about a given environment.}

\subsection{Summary Type Differences}

Considering the results in Tab. \ref{tab:summary-type-evaluation}, but also the baseline results in Tab. \ref{tab:baseline-similarities} and \ref{tab:baseline-vision-similarities}, \emph{RQ3 (summarisation using natural language or temporal logic) cannot be answered conclusively, as the similarity with the ground-truth summaries varies between skills and is not consistently better for either summary type}.
For the majority of skills and models, the LTL summary seemingly leads to higher similarity with the ground truth; however, this is only consistently the case for \emph{Gemma 4 31B}, while the differences for the other models are not always prominent.
This indicates that, without finetuning a model for a specific summary type, comparable semantic envelopes can be expected with both summary types.

By qualitatively analysing the generated LTL formulas, we could also notice that, while semantically close to the ground truth, the generated LTL formulas do not always capture the nuances of the execution process entirely correctly.
For instance, the LTL-based semantic envelope extracted by \emph{Gemma 4 31B} for the \emph{cup pickup} skill is $F(\texttt{inFrontOf}(robot, cup)) \land F(\texttt{holding}(robot, cup))$; as shown in Tab. \ref{tab:summary-type-evaluation}, the semantic similarity with the ground truth is particularly high in this case.
While technically correct, a version that would capture the execution better may be $F(\texttt{inFrontOf}(robot, cup)) \; U \; \texttt{holding}(robot, cup)$, as this would constrain the motion of the robot throughout the execution.
This example illustrates that the semantic similarities for the translated LTL formulas are potentially higher than they should be, which further supports the claim that we cannot provide a conclusive answer to RQ3 based on the obtained results.

\subsection{Sensitivity to Vocabulary Changes}

With respect to RQ2, another aspect we investigate is how the generated summaries $\vec{m}_i^j$ are affected by changes in the vocabulary $V$.
For this, we regenerate the summaries using the following reduced vocabulary $V^-$, which uses about $60\%$ of the relations in the original $V$:
\begin{mdframed}[style=roundframe]
    \vspace{-0.4cm}
    \begin{equation*}
        \begin{split}
            V^- = \{ &\texttt{leftOf}(x, y), \texttt{rightOf}(x, y), \texttt{above}(x, y), \texttt{below}(x, y), \texttt{inFrontOf}(x, y), \\
                     &\texttt{behind}(x, y), \texttt{nextTo}(x, y), \texttt{on}(x, y), \texttt{in}(x, y), \texttt{holding}(x, y) \}
        \end{split}
    \end{equation*}
\end{mdframed}
We then perform Welch's t-test between the similarities in Tab. \ref{tab:summary-type-evaluation} and the similarities computed for the regenerated summaries.

Tab. \ref{tab:vocabulary-change-sensitivity} shows the averaged similarities with the ground-truth summaries for all skills and models, generated with $V^-$, along with results from the t-test.
\begin{table}[t]
    \centering
    \scriptsize
    \caption{Average semantic similarities with the ground-truth summaries (over all skill episodes) after decreasing the vocabulary. A $*$ denotes a statistically significant difference from the similarities in Tab. \ref{tab:summary-type-evaluation} at $\alpha=0.05$, and a $**$ a statistically significant difference at $\alpha=0.01$.}
    \label{tab:vocabulary-change-sensitivity}
    \begin{tabular}{M{0.12\linewidth} | M{0.135\linewidth} | M{0.135\linewidth} | M{0.135\linewidth} | M{0.135\linewidth} | M{0.135\linewidth} | M{0.135\linewidth}}
        \cellcolor{blue!10} & \multicolumn{2}{c}{\cellcolor{blue!10}\textbf{Claude Opus 4.6}} & \multicolumn{2}{c}{\cellcolor{blue!10}\textbf{DeepSeek V4 Flash}} & \multicolumn{2}{c}{\cellcolor{blue!10}\textbf{Gemma 4 31B}} \\\cline{2-7}
        \multirow{-2}{*}{\cellcolor{blue!10}\textbf{Skill}} & \cellcolor{blue!10}\textbf{Language summary} & \cellcolor{blue!10}\textbf{LTL formula} & \cellcolor{blue!10}\textbf{Language summary} & \cellcolor{blue!10}\textbf{LTL formula} & \cellcolor{blue!10}\textbf{Language summary} & \cellcolor{blue!10}\textbf{LTL formula} \\\hline
        \cellcolor{blue!10}Cup pickup     & $0.85 \pm 0.02$ & $0.94 \pm 0.01$ & $0.90 \pm 0.04$ & $0.87 \pm 0.05$ & $0.82 \pm 0.06^{*}$ & $0.96 \pm 0.02$ \\\hline
        \cellcolor{blue!10}Cup placing    & $0.87 \pm 0.03$ & $0.85 \pm 0.03^{**}$ & $0.79 \pm 0.05^{*}$ & $0.84 \pm 0.02$ & $0.87 \pm 0.06$ & $0.90 \pm 0.03$ \\\hline
        \cellcolor{blue!10}Cup moving     & $0.85 \pm 0.01$ & $0.82 \pm 0.11$ & $0.80 \pm 0.05^{**}$ & $0.80 \pm 0.12$ & $0.80 \pm 0.04$ & $0.88 \pm 0.02^{**}$ \\\hline
        \cellcolor{blue!10}Spoon picking  & $0.84 \pm 0.02$ & $0.90 \pm 0.02$ & $0.89 \pm 0.02$ & $0.88 \pm 0.07$ & $0.86 \pm 0.03$ & $0.92 \pm 0.02$ \\\hline
        \cellcolor{blue!10}Spoon placing  & $0.92 \pm 0.01$ & $0.93 \pm 0.03$ & $0.85 \pm 0.06$ & $0.92 \pm 0.02$ & $0.83 \pm 0.05$ & $0.94 \pm 0.02$ \\\hline
    \end{tabular}
\end{table}
As can be seen, the vocabulary reduction does seem to affect the summaries significantly, such that, compared to Tab. \ref{tab:summary-type-evaluation}, the similarities with the ground-truth descriptions are generally slightly lower than when the original $V$ is used; however, except for a few cases, these differences are not statistically significant, which indicates that the models can generally deal with minor changes in vocabulary.

\subsection{Applicability for Execution Monitoring}
\label{sec:evaluation:monitoring}

\begin{figure}[t]
    \centering
    \includegraphics[width=\linewidth]{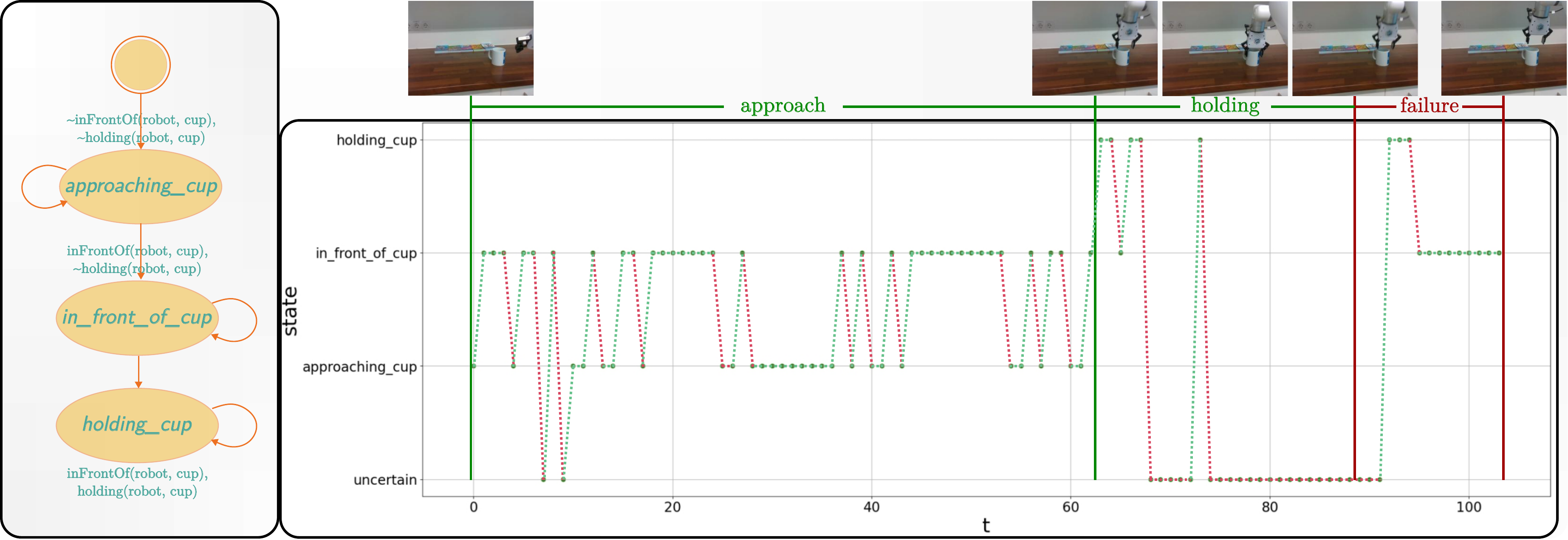}
    \caption{A monitoring example for a failed cup pickup episode using the generated graph $\mathcal{G}_{\skill_i}$ shown in the left part of the figure (monitoring done using \emph{Claude Opus 4.6}). The timeline above the graph is a manual, ground-truth annotation of the execution stages just for illustration. Each point in the plot represents the identified execution state at each time $t$, green dotted lines denote valid transitions according to the monitoring graph, while red dotted lines denote invalid transitions. The failure in this case is that the robot drops the cup after grasping it. While the robot is holding the cup, the model seems to be affected by a subtle occlusion caused by the robot and is uncertain about the state most of the time; however, once the failure occurs and after a slight delay, there is correct recognition that the robot is back to the \emph{in\_front\_of\_cup} state.}
    \label{fig:annotated-monitoring-pipeline-cup-pickup}
\end{figure}

We study RQ4 (the benefit of semantic descriptions for execution monitoring) qualitatively by illustrating the feasibility of using $\mathcal{G}_{\skill_i}$ for execution state monitoring.
Fig. \ref{fig:annotated-monitoring-pipeline-cup-pickup} shows a monitoring timeline for one of the anomalous pickup episodes, where we used \emph{Claude Opus 4.6} to identify the state $N_{\skill_i,t}$.
The monitoring graph $\mathcal{G}_{\skill_i}$ for the cup pickup skill, also shown in the figure, was generated using \emph{Claude} based on the semantic envelope given in the form of an LTL formula, also generated by \emph{Claude}.
For ease of interpretability, we have annotated the timeline to show the key frames in the execution.
This figure demonstrates that $\mathcal{G}_{\skill_i}$ can be useful for monitoring, as the different semantic stages of the execution process (approaching, grasping, and then the failed grasp) can be clearly identified from the monitoring timeline; however, it also illustrates that pretrained models are easily affected by scene occlusions, which may cause incorrect state recognition (in this case, the occlusion is very subtle and is caused by the gripper covering parts of the cup as it grasps it).
In fact, it should be noted that Alg. \ref{alg:monitoring} would raise frequent alarms in the annotated timeline in Fig. \ref{fig:annotated-monitoring-pipeline-cup-pickup} (illustrated by the red dotted lines that indicate invalid transitions); this is because the monitoring graph in Fig. \ref{fig:annotated-monitoring-pipeline-cup-pickup} does not include a transition from the \emph{in\_front\_of\_cup} to the \emph{approaching\_cup} state.
This also illustrates a strength in our approach, however: as the generated monitoring graph is easy to interpret, it could easily be integrated in a human-in-the-loop scenario where the user could add the missing transition to the graph and thus avoid false alarms.\footnote{While this is beyond the scope of the work, such modifications could be done through a basic graphical user interface, which would be simple to implement because the generated graphs are represented as JSON dictionaries.}
The VLM itself also plays an important role here.
For instance, we also attempted using \emph{SpaceOm} for monitoring in the same pickup episode for comparison, but we noticed that the model often recognises the \emph{holding\_cup} state during the approach stage; such incorrect detections would lead to a significant number of false alarms that cannot be resolved by improving the graph itself, but would also require model finetuning for better state recognition.

\begin{figure}[t]
    \centering
    \includegraphics[width=\linewidth]{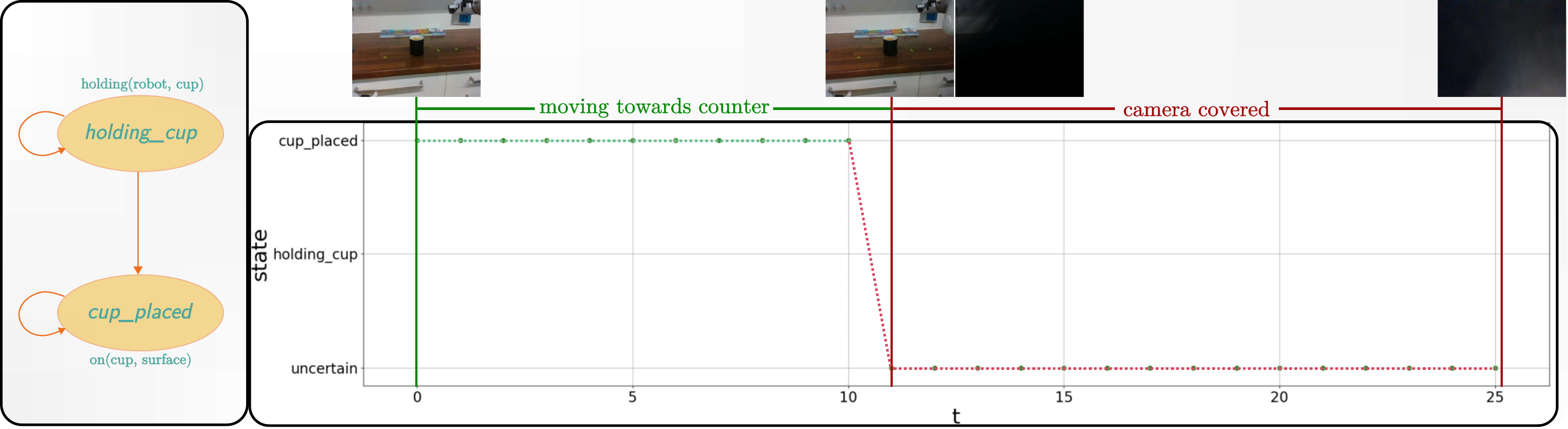}
    \caption{A monitoring example for an anomalous cup placing episode using the generated graph $\mathcal{G}_{\skill_i}$ shown in the left part of the figure (monitoring done using \emph{Claude Opus 4.6}). As before, the timeline above the graph is a manual, ground-truth annotation of the execution stages. Each point in the plot represents the identified execution state at each time $t$, green dotted lines denote valid transitions according to the monitoring graph, while red dotted lines denote invalid transitions. In this case, the camera gets covered during the execution and remains covered until the end of the episode. While this stage of the execution is correctly identified as an uncertain state, the first stage of the execution is classified as a \emph{cup\_placed} state; this is likely because the model focuses on the cup that is already on the counter rather than the one that needs to be placed by the robot.}
    \label{fig:annotated-monitoring-pipeline-cup-placing}
\end{figure}

We show another example of a monitoring timeline for an anomalous cup placing episode in Fig. \ref{fig:annotated-monitoring-pipeline-cup-placing}.
The monitoring graph used in the case was generated from a semantic envelope given in the form of an LTL formula generated by \emph{Gemma}; this aims to demonstrate that the ability to generate useful graphs does not have a strong dependency on the semantic envelopes generated by a specific model.\footnote{It can be noted that, unlike the graph for the cup pickup case, this graph does not have an initial state. This is not a Gemma-specific phenomenon, as we also noticed it for the other models for some of the skills. For more information, an interested reader can look through the data of the generated graphs for all skills, which are included as part of our code repository.}
As above, we used \emph{Claude Opus 4.6} to identify the state at each step of the execution.
This example illustrates the correct handling of another type of failure (a covered camera), but also shows another state recognition problem, where the model focuses on an incorrect object in the scene.
Problems of this type cannot be resolved by making modifications to the monitoring graph, but could potentially also be mitigated by providing a more detailed task specification $L_i$.

\subsection{Generalisation to Another Robot}

The final aspect that we illustrate is the ability to generalise $\mathcal{G}_{\skill_i}$, both over objects and to other robots.
We concretely consider a case of another robot (a PAL TIAGo) with the task of grasping an apple.
To assess the generalisation of $\mathcal{G}_{\skill_i}$, we use the same graph shown in Fig. \ref{fig:annotated-monitoring-pipeline-cup-pickup} for monitoring this new execution; for this example, the updated task description $L_i$ is \emph{Pick the apple}, and all instances of \emph{cup} are grounded to \emph{apple}.
Here, it should be noted that, as TIAGo has a different physical embodiment, its view of the scene is different than for the robot used in the COFFAIL dataset (from which $\mathcal{G}_{\skill_i}$ was originally extracted).

\begin{figure}[t]
    \centering
    \includegraphics[width=\linewidth]{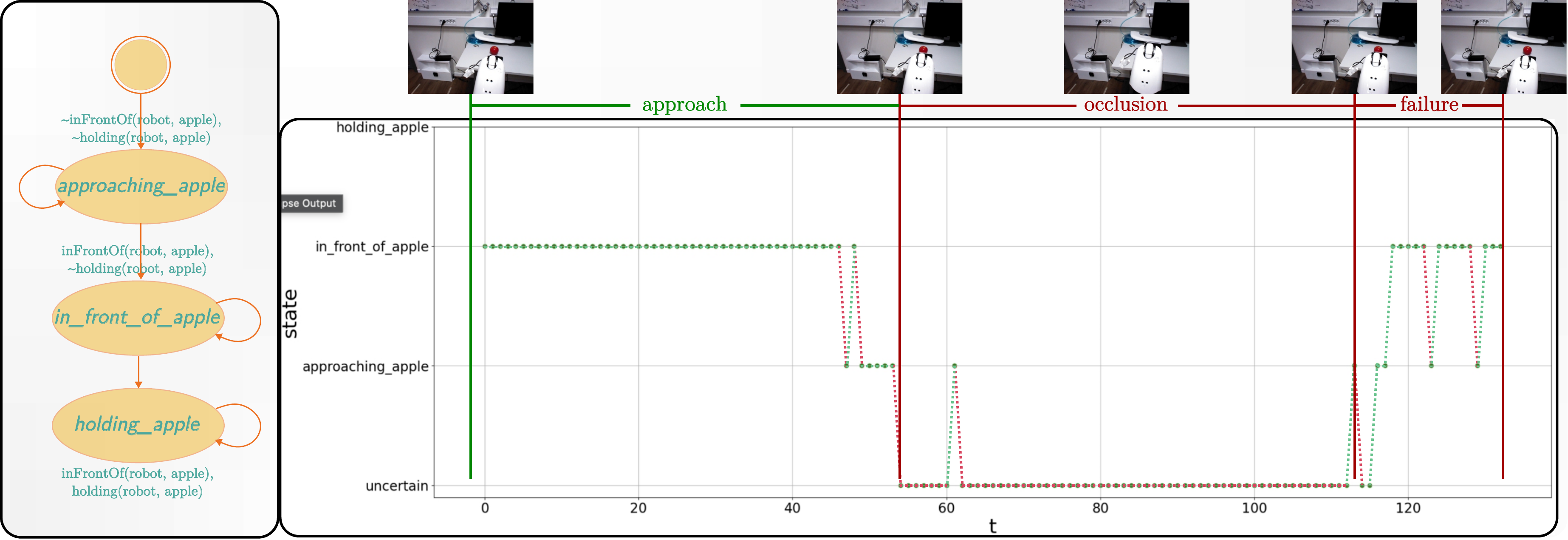}
    \caption{A monitoring example for a failed apple pickup execution episode with a PAL TIAGo robot, using the generalised $\mathcal{G}_{\skill_i}$ from Fig. \ref{fig:annotated-monitoring-pipeline-cup-pickup} (monitoring done using \emph{Claude Opus 4.6}). The timeline above the graph is once again a manual, ground-truth annotation of the execution stages for illustration purposes. Each point in the plot represents the identified execution state at each time $t$. The failure here is that the robot does not manage to grasp the object. During the approach phase, the robot occludes the object, so the model is uncertain about the state. Once the object is visible again, there is correct recognition that the robot is back to the \emph{in\_front\_of\_apple} state, or equivalently, that the object has not been grasped.}
    \label{fig:annotated-monitoring-pipeline-apple-pickup-tiago}
\end{figure}

Fig. \ref{fig:annotated-monitoring-pipeline-apple-pickup-tiago} shows a monitoring timeline for an anomalous grasping episode with TIAGo, where the robot occludes the apple during the approach phase and then fails to grasp it.
This example demonstrates that it is possible to transfer the generalised $\mathcal{G}_{\skill_i}$ for state monitoring between robots.
In fact, this illustrates one of the main strengths of $\mathcal{G}_{\skill_i}$: as the graph encodes abstracted information about the execution process, it is unaffected by the fact that the scene view is different from the one from which $\mathcal{G}_{\skill_i}$ was extracted (although this may affect the monitoring VLM itself).

%% file: subfiles/discussion.tex
\section{Discussion and Conclusion}
\label{sec:discussion}

\paragraph{Conclusion}
In this paper, we presented a combined VLM-LLM method to extract an interpretable semantic envelope, which is a description of constraints (either in natural language or as an LTL sentence) that should hold during or after the execution of a given robot skill.
Our method processes nominal execution episodes, such that it uses a VLM to annotate each execution step based on a predefined vocabulary, which is represented as a set of lifted predicates; these annotations are then used in an LLM to extract an episode-specific constraint description.
The semantic envelope is extracted by combining the episode-specific descriptions; from this, we create a graph of execution states that can be used for execution monitoring.
On a dataset of various everyday skills, we showed that the inclusion of observational evidence improves upon baseline semantic envelopes (generated without any or with minimal execution evidence), and demonstrated high cosine similarities (generally higher than $0.85$ across skills) between the generated envelopes and ground-truth descriptions.
We additionally demonstrated the feasibility of using the execution graph for monitoring skill execution, which allows a robot to identify failures such as failed grasps and camera occlusions, as well as that the monitoring graph can be transferred between robots.

Overall, our work shows that open-world models can be used for describing robot execution semantically and for execution monitoring based on such descriptions, provided that the complete process is guided by structured information and carefully-crafted prompts with embedded constraints that together produce meaningful execution-specific output.

\paragraph{Discussion and future work}
There are various aspects that future work could consider.
In this work, our objective was to investigate the overall ability of VLMs and LLMs to produce a meaningful summary of a skill's execution process in terms of a predefined vocabulary, such that we evaluated the correctness of generated descriptions through a comparison with manually crafted ground-truth descriptions of execution constraints.
It would, however, also be worthwhile to explore the integration of methods that can automatically fix generated formulas, such as \cite{li2026-2}, to ensure more general, formal correctness.
Furthermore, as mentioned when discussing the monitoring examples, the actual usefulness of the generated graphs for monitoring depends on the ability to recognise execution states accurately.
Using large pretrained (and thus open-world) models, as we have done in this work, has the benefit of not being constrained to recognising a predefined number or types of states, but has the downside of difficult-to-predict overall accuracy; particularly in safety-critical applications, it may be more sensible to use dedicated state classifiers, such as in \cite{sliwowski2025}.
In this work, we manually defined a vocabulary from which execution descriptions are extracted; this is in line with the general literature on task and motion planning, but is also a deliberate design choice, as the method can be easily adapted to use domain-specific vocabularies, which is particularly useful in human-in-the-loop or collaborative scenarios, where a robot's operation needs to be closely monitored by a human operator.
Nevertheless, it would also be useful to investigate extensions of the framework to use derived symbols, for instance as in \cite{ahmetoglu2025,shao2026}, though this would require a more visual and less textual representation of the execution process.
Finally, our purpose of developing a semantic representation for monitoring is not only because we are interested in post-hoc failure detection, but also because we want to embed the monitoring process into a predictive execution loop for failure prevention; this is an aspect we are considering in our ongoing work.

%% file: subfiles/appendix.tex
\section*{Appendix: Qualitative Results and Important Design Notes}
\label{sec:appendix}

\begin{figure}[h]
    \centering
    \begin{subfigure}[t]{0.23\linewidth}
        \includegraphics[width=\linewidth]{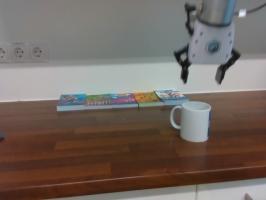}
        \caption{\emph{The robot is in front of the cup and appears to be moving towards it, possibly preparing to pick it up.}}
    \end{subfigure}
    \hspace{0.01\linewidth}
    \begin{subfigure}[t]{0.23\linewidth}
        \includegraphics[width=\linewidth]{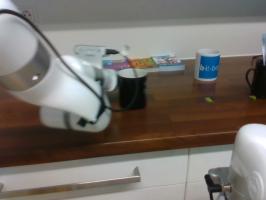}
        \caption{\emph{The robot is pushing the cup forward.}}
    \end{subfigure}
    \hspace{0.01\linewidth}
    \begin{subfigure}[t]{0.23\linewidth}
        \includegraphics[width=\linewidth]{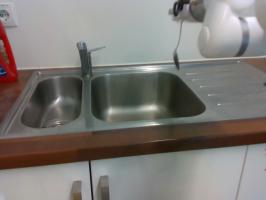}
        \caption{\emph{The robot is holding the spoon in its hand and is positioned in front of the sink, likely preparing to place the spoon inside.}}
    \end{subfigure}
    \hspace{0.01\linewidth}
    \begin{subfigure}[t]{0.23\linewidth}
        \includegraphics[width=\linewidth]{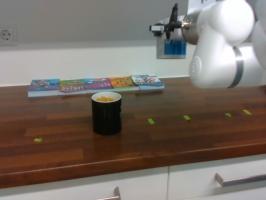}
        \caption{\emph{The robot is holding the cup and moving it towards the table or another surface where it intends to place the cup.}}
    \end{subfigure}
    \caption{Illustrations of generated episode step annotations on representative dataset examples.}
    \label{fig:episode-step-annotations}
\end{figure}

To simplify the interpretation of the quantitative analysis presented in Sec. \ref{sec:evaluation}, we also present various qualitative results.
In Fig. \ref{fig:episode-step-annotations}, we illustrate various execution step descriptions $D^j_{i,t}$ generated by \emph{SpaceOm}, which show that the captions generally capture relevant information about the scene.

\begin{figure}[h]
    \centering
    \begin{subfigure}{0.19\linewidth}
        \includegraphics[width=\linewidth]{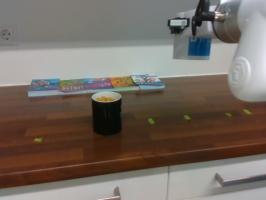}
    \end{subfigure}
    \begin{subfigure}{0.19\linewidth}
        \includegraphics[width=\linewidth]{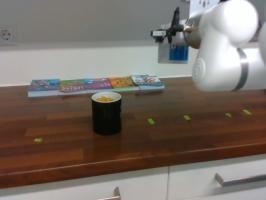}
    \end{subfigure}
    \begin{subfigure}{0.19\linewidth}
        \includegraphics[width=\linewidth]{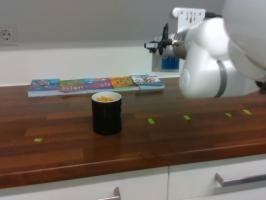}
    \end{subfigure}
    \begin{subfigure}{0.19\linewidth}
        \includegraphics[width=\linewidth]{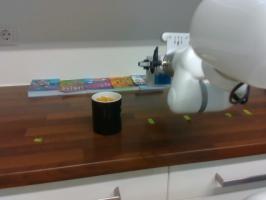}
    \end{subfigure}
    \begin{subfigure}{0.19\linewidth}
        \includegraphics[width=\linewidth]{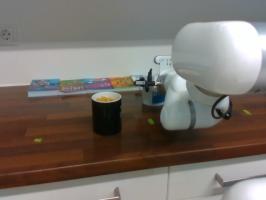}
    \end{subfigure}
    \caption{Illustration of a generated summary for one of the cup placing episodes: \emph{During the execution of 'place the cup you are holding', the robot should be \texttt{holding}(robot, cup) and the cup should be \texttt{above}(cup, surface), with the robot positioned \texttt{inFrontOf}(robot, surface) while moving the cup toward its target placement location.}}
    \label{fig:episode-summary-illustration}
\end{figure}

\noindent Fig. \ref{fig:episode-summary-illustration} shows one of the cup placing episodes and the generated episode summary $\vec{m}^j_i$.

\noindent Finally, in Tab. \ref{tab:generated-envelopes}, we show the generated semantic envelopes for all skills by \emph{Claude Opus 4.6}, along with the similarities with the ground-truth summaries.
\begin{table}[t]
    \centering
    \scriptsize
    \caption{Generated semantic envelopes $\mathcal{E}_{\skill_i}$ for the skills of interest using \emph{Claude Opus 4.6}}
    \label{tab:generated-envelopes}
    \begin{tabular}{M{0.065\linewidth} | M{0.2\linewidth} | M{0.11\linewidth} | M{0.25\linewidth} | M{0.2\linewidth} | M{0.11\linewidth}}
        \cellcolor{blue!10}\textbf{Skill} & \cellcolor{blue!10}\textbf{Language envelope} & \cellcolor{blue!10}$\operatorname{sim}(\vec{g}_i, \mathcal{E}_{\skill_i})$ & \cellcolor{blue!10}\textbf{LTL envelope} & \cellcolor{blue!10}\textbf{Translated LTL envelope} & \cellcolor{blue!10}$\operatorname{sim}(\vec{g}_i, \mathcal{E}_{\skill_i})$ \\\hline
        \cellcolor{blue!10}Cup pickup     & \emph{During the execution of 'pick the cup', the robot must be inFrontOf(robot, cup) and eventually holding(robot, cup).} & $0.87$ & \tiny $\begin{array}{l}F(inFrontOf(robot, cup) \\U \; holding(robot, cup))\end{array}$ & \emph{Eventually, the robot will be in front of the cup and will remain in front of it until it is holding the cup.} & $0.93$ \\\hline
        \cellcolor{blue!10}Cup placing    & \emph{During the execution of 'place the cup you are holding', the robot should be holding(robot, cup) and the cup should be above(cup, surface), with the robot positioned inFrontOf(robot, surface) while moving the cup toward its target placement location.} & $0.90$ & \tiny $\begin{array}{l}holding(robot, cup) \\U (on(cup, table) \\\lor on(cup, counter))\end{array}$ & \emph{The robot continues holding the cup until it is placed on either the table or the counter.} & $0.81$ \\\hline
        \cellcolor{blue!10}Cup moving     & \emph{During the execution of 'move the cup', the robot should be pushing(robot, cup) or holding(robot, cup).} & $0.87$ & \tiny $G(pushing(robot, cup) \lor holding(robot, cup))$ & \emph{At all times, the robot must either be pushing or holding the cup.} & $0.90$ \\\hline
        \cellcolor{blue!10}Spoon picking  & \emph{During the execution of 'pick up the spoon', the robot should be inFrontOf(robot, spoon) and holding(robot, spoon).} & $0.86$ & \tiny $F(inFrontOf(robot, spoon) \land F(holding(robot, spoon)))$ & \emph{Eventually, the robot will be in front of the spoon, and after that, it will eventually be holding the spoon.} & $0.90$ \\\hline
        \cellcolor{blue!10}Spoon placing  & \emph{During the execution of placing the spoon in the sink, the robot should be holding(robot, spoon) and inFrontOf(robot, sink), with the spoon above(spoon, sink).} & $0.93$ & \tiny $\begin{array}{l}holding(robot, spoon) \; \land \\inFrontOf(robot, sink) \; U\\(holding(robot, spoon) \; \land \\above(spoon, sink) \; \land \\inFrontOf(robot, sink)) \\U in(spoon, sink)\end{array}$ & \emph{The robot must first be holding the spoon while standing in front of the sink, then continue holding the spoon while positioning it above the sink and remaining in front of it, until finally the spoon is placed inside the sink.} & $0.96$
    \end{tabular}
\end{table}

One relevant aspect to mention is that it appears to be important for the episode step annotations $D^j_{i,t}$ to be expressed in natural language.
In particular, we also experimented with the descriptions being expressed as conjunctions of ground predicates from $V$; however, in this case, we noticed that the grounding was too inconsistent, both in terms of the constant names (e.g. \emph{robot} on one occasion, but \emph{robotHand} on another) and in terms of the constant ordering (e.g. \texttt{in}(cup, robot) on one occasion, but \texttt{in}(robot, cup) on another).
With language-based summaries, such inconsistency did not seem to propagate as much, as grounding is only done when the episode summaries and envelopes are generated.

One final important aspect to note is that, considering the use of VLMs and LLMs throughout our method, all elements of our framework (step annotation, episode summary extraction, semantic envelope generation, monitoring graph generation, and monitoring) are clearly sensitive to the provided prompts. The prompts and all constraints within them, which we included as part of this paper, were designed in a careful trial-and-error process so that they produce reasonable outputs and can be transferred between different models.